\documentclass[preprint,12pt,number]{elsarticle}

\usepackage{graphicx,amssymb,amsmath,amsfonts,subfigure, algorithm, algorithmic, multirow, amsmath, xcolor, graphics, graphicx, epsfig, multirow,bm,epsfig}
\journal{Information Sciences}

\begin{document}

\begin{frontmatter}



\title{Spatio-Temporal Road Scene Reconstruction Using Superpixel Markov Random Field}

 \author[label1]{Yaochen Li}
 \author[label2]{Yuehu Liu}
 \author[label1]{Jihua Zhu}
 \author[label3]{Shiqi Ma}
 \author[label2]{Zhenning Niu}
 \author[label1]{Rui Guo}
 \address[label1]{School of Software Engineering, Xi'an Jiaotong University}
 \address[label2]{Institute of Artificial Intelligence and Robotics, Xi'an Jiaotong University}
 \address[label3]{School of Information and Electronics, Beijing Institute of Technology}

%\author{}

%\address{}

\begin{abstract}
Scene model construction based on image rendering is an indispensable but challenging technique in computer vision and intelligent transportation systems.
In this paper, we propose a framework for constructing 3D corridor-based road scene models.
This consists of two successive stages: road detection and scene construction.
The road detection is realized by a new superpixel Markov random field (MRF) algorithm.
The data fidelity term in the MRF's energy function is jointly computed according to the superpixel features of color, texture and location.
The smoothness term  is established on the basis of the interaction of spatio-temporally adjacent superpixels.
In the subsequent scene construction, the foreground and background regions are modeled independently.
Experiments for road detection demonstrate the proposed method outperforms the state-of-the-art in both accuracy and speed.
The scene construction experiments confirm that the proposed scene models show better correctness ratios, and have the potential to support a range of applications.
%Useful applications can be developed based on the proposed framework, such as traffic simulation and virtual street view.
\end{abstract}


\end{frontmatter}


\section{Introduction}
\label{Intro}
With the rapid development of information science and computer vision, scene construction based on image rendering has become an important feature of many  applications.
Scene construction results are of particular use when seeking to simulate real-world events.
There are three kinds of applications typically associated with scene construction:
%(1) High-level vision tasks.
%Scene reconstruction is the precondition of many high-level vision tasks, such as action recognition, scene understanding, etc.
(1) Virtual street views \citep{Google_View,Street_Slide}.
Road image sequences contain lots of  information people may find useful.
In response to this, Google sent out a fleet of cars to capture road images from all over the world.
The Google Street View system was developed to allow users to tour street scenes while browsing a map.
Another similar application is Microsoft Street Slide, which introduces scene bubbles that users can  interact with.
(2) Intelligent transportation systems \cite{Spatio_Temporal_Traffic, Cognitive_Cars}.
Advanced driver assistant systems (ADAS) are be improved through virtual scene reconstruction.
In contrast to traditional field tests of unmanned vehicles, an evaluation can be implemented in a virtual environment.
The simulation-based evaluation system can handle thousands of quantitative judgments in a short time and be more objective than a human expert \cite{Science_Li}.
%The off-line test for unmanned vehicles is much safer and saves time and energy.
(3) Free viewpoint television \cite{Virtual_Video-Camera, Free_Viewpoint_TV}.
Virtual reality technology  is commonly used in the modern film industry.
Free viewpoint video has become popular because, it provides users with richer interactions.
Users can freely select their viewpoint while browsing  videos by drawing upon free viewpoint rendering methods.\par
%Moreover, users can change their viewpoint freely by adopting the free-viewpoint rendering technology

In early studies, virtual scenes were constructed by using computer graphics approaches.
Typical examples include the Prescan software \footnote{http://www.tassinternational.com/prescan}, and IPG Co.'s CarMaker software \cite{CarMaker}.
Alongside of computer graphics methods, road scene construction from image sequences became popular.
The Waymo team at Google proposed a virtual platform for scene modeling from road images, utilizing machine learning methods, with off-line testing of unmanned vehicles being conducted using the platform.
%The off-line test of unmanned vehicles is conducted in the platform.
Recently, generative adversarial networks (GAN) have been proposed that can synthesize virtual traffic scenes \cite{GAN}.
However, all of these methods have a variety of limitations, including issues with modeling both the foreground and background, rendering of new viewpoint images, and so on.

%In this theory, the purpose of vision computing  to recover the 3D shapes of objects in the scene.
Traditional reconstruction of road scene methods do not distinguish between the foreground and background regions, such as the framework of Google Street View \cite{Google_View} and Microsoft Street Slide \cite{Street_Slide}.
Tour into the picture (TIP) framework models both the foreground and background regions of an image \cite{Tour_Into_Picture, Tour_Into_Video, T_ITS}.
After the specification of the vanishing point, the background model is roughly composed of ``left wall'', ``right wall'', ``back wall'', ``sky'' and ``road plane''.
%After the detection of the vanishing point, the input image is roughly partitioned into ``left wall'', ``right wall'', ``back wall'' and ``road plane''.
The foreground models can be constructed independently of the background models.
However, the TIP model is not fit for the curved-edge road conditions.
%The foreground polygons are assumed to stand vertically on the ``floor'' plane of the background model.
In this paper, a new framework to model 3D scenes is introduced using David Marr's theory \cite{David_Marr}, which builds upon the vision framework of ``feature map computation-depth recovery-3D modeling''.
%The idea of the proposed framework is similar to the TIP model
%The proposed framework builds upon the idea of TIP model,
%The vision system of ``component generation-depth computation-3D recovery'' was considered a classical model in computer vision community.
%Besides the reconstruction of semantic objects, the semantic analysis and modeling of the entire scene becomes an important research problem.
The idea of the TIP model is also incorporated, which implements the region detection and semantic reasoning to automatically construct the 3D road scenes.
%The semantic analysis of the road scenes is an important preliminary.
%In order to implement scene analysis and modeling, the region detection and semantic reasoning are crucially important.
%The 3D scene structure reconstruction is then implemented based on these technologies.
%The purpose of region detection is to decompose an image into meaningful regions, and then category these regions into object representations.
%In this paper, a new framework to model 3D scenes is introduced using David Marr's theory \cite{David_Marr}, 
The road regions of an input image sequence are detected  via a new superpixel-based Markov random field (MRF) method, with each superpixel refering to a group of connected pixels of a similar color.
An energy function is defined that is  based on vision features computed in the spatio-temporal domain.
An energy minimization process is conducted following a cycle of ``global energy initialization-local energy comparison-global energy comparison''.
After the detection of road regions, 3D corridor-structured scene models are constructed, based on road boundary control points.
The road regions are assumed to be on a horizontal plane, while the rest of the scene components are treated as  standing perpendicularly to the road plane.
Panoramic scene models can also be constructed, and it can serve as the basis for a range of potential applications, such as virtual street view, traffic scene simulations, etc.\par

The main contributions of this paper are threefold:
   \begin{itemize}
   \item A new superpixel-based MRF method that can detect road regions in image sequences.
    %A superpixel-based MRF algorithm is designed for energy minimization.
    The data fidelity term in the MRF's energy function is defined according to a combination of superpixel features of color, texture and location.
    Its smoothness term is computed according to the feature distance between of spatio-temporally adjacent superpixels.
    Energy minimization is implemented by means of a ``global energy initialization-local energy comparison-global energy comparison'' cycle.\par

   \item %The spatio-temporally corresponded scene models are proposed for road image sequences.
         A novel framework to construct road scene models automatically from image sequences.
         The road scene model has a 3D corridor structure, with the road regions being assumed to be  horizontal planes.
         Foreground objects are modeled independently of the background models.
         Panoramic scene models can also be constructed.

   \item A new intelligent system for  interactive tour in traffic scenes, based on the proposed scene models.
         Two basic modes are designed: bird's eye view mode; and touring mode.
         New traffic elements can be inserted into the traffic scenes, which can be used for  virtual simulations with unmanned vehicles.
%Based on the road detection results, the control points of road boundaries are computed with the constraint of vanishing points.
%The control points are utilized as the basis for 3D scene models construction.

\end{itemize}





\section{Related Works}
The research presented in this paper focuses on road detection via superpixel-based MRF, and on scene model reconstruction.
%can be decomposed into two parts: (1) road region detection and (2) scene models construction.
Below, we discuss the related research for each of these components, respectively:\par

\textbf{(1) Region detection:}
The semantic analysis of road images is an important part of 3D road scene reconstruction.
Road detection underpins this semantic analysis.
Belaid et al.  have proposed a watershed transformation method, that can be applied to road region detection \cite{Watershed}.
A topological gradient method is used to avoid over segmentation.
%Jiang et al. \cite{Hanqing_Jiang} propose a method to extract spatio-temporally consistent segments from a video.
%The corresponding depth data is estimated and 3D reconstruction is implemented.
% by the multi-view stereo technique.
%The applications of 3D reconstruction, video editing and semantic segmentation are also developed.
Chen et al. \cite {IS_Chen} proposed a mechanism to support the long-term background and short-term foreground models.
A unified Bayesian framework is utilized in the background-foreground fusion stage.
Song et al. \cite{Zhichao_Song} have propose a new method for image appearance transfer that combines color and texture features.
To do this, they imlemented feature detection and matching between source and reference images.
%For road regions detection, Alvarez et al. \cite{Alvarez} combine road priors and contextual information to detect the road regions.
%The contextual cues mainly include horizon lines, vanishing points, lane markings, etc.
%Inspired by dynamic programming, Yao et al. \cite{Yao_J} reduce the free space estimation task to an inference problem on a 1D graph.
The image and geometric features were then able to be exploited.
%Peng et al. \cite{Jianteng_Peng} propose a method to extract superpixels using a higher order optimization framework.
%The K-means clustering technique is  adopted, and a higher order energy function is employed to optimize the initial superpixels.
%The higher order energy method is also used in the study of \cite{Jianbing_Shen}.
%Dong et al. [Xingping Dong] propose a sub-Markov random walk algorithm with label prior for seeded image segmentation.
%The algorithm is considered as a traditional random walker on a graph with auxiliary nodes.\par
%Besides the image segmentation, video salient object detection is also an important research topic [Wenguan Wang].\par
%In this paper, we propose a novel superpixel MRF method for road region detection of video sequences.
Xiao et al. \cite{Liang_Xiao} have sought to  extend the traditional conditional random field (CRF) model by proposing a hybrid CRF model that integrates the camera information   and LIDAR.
The hybrid CRF model can be optimized with graph cuts to acquire road regions.
However, there is a dependency upon  LIDAR information for this approach to work.
Kim et al. \cite{Chansu_Kim} have introduced a search method based on a coarse-to-fine strategy and image superpixels.
A simple generative appearance model is applied during the initial coarse step.
In the refine step, a sampling and similarity measurement process is performed.
Gould et al. \cite{Graph_Label} have presented a semantic segmentation method that uses label transfer.
An approximate nearest neighbor algorithm is applied to build a graph upon the superpixels.
Although this approach is effective for standard datasets, experiments using road image sequences have not always been satisfactory.
Lu et al. \cite{GrowCut} used an unsupervised method to select superpixel neighbors according to the GrowCut framework.
However, for this to work the vanishing points have to be detected first, as a supplementary step.

\textbf{(2) Superpixel-based MRF:}
Pixel-level MRF algorithms are widely used for image segmentation and annotation.
Kim et al. \cite{Kim_IY} have applied an MRF model to image segmentation, that is especially effective for outdoor images.
Yang et al. \cite{Yang_F} have  proposed a pixon-based MRF method for image segmentation.
The basic idea of a ``pixon'' is that the spatial scale at each image site will vary according to the image information.
%Less computation cost of the energy optimization is demanded than the traditional MRF algorithms.
However, the generation of pixons requires the solving of an anisotropic diffusion equation.
%As a result, the segmentation results are usually irregular and uncertain.
Elia et al. \cite{Elia_CD} have proposed a tree-structured MRF model, that can be applied to Bayesian image segmentation.
Here, the input image can be iteratively segmented into smaller regions based on a series of ``split-merge'' steps.
However, pixel-based MRF methods are very susceptible to noise.
Schick et al. \cite{Schick_A} have proposed a superpixel-based MRF algorithm, that uses a classical Graphcut approach to energy minimization.
The graph cut approach aims to solve the segmentation problem using max flow and min cut theory.
However, the data fidelity term in the energy function is too simple to be able to reflect the irregular attributes of superpixels.
%Pei et al. \cite{Pei_SC} propose a belief propagation method to detect the image regions based on superpixel saliency.
Fulkerson et al. \cite{Superpixel_CRF} have constructed a classifier by aggregating histograms of superpixel neighborhoods.
A superpixel is a group of connected pixels with similar color information, which is a polygonal part of an image.
%This method firstly partition the input image into mid-level superpixels.
The superpixel graph is then optimized by a CRF mechanism.
%The vision features of superpixels are extracted to optimize the saliency regions.
However, superpixel features are insufficient in this method, and the spatio-temporal superpixel relationships are ignored.
A variety of methods is proposed in \cite{Wang_XF}, but these also have drawbacks relating to the insufficient use of temporal information, complex iterations of the energy function and low efficiency  in the data fidelity terms.
To overcome the drawbacks and limitations mentioned above, we are proposing here a new superpixel-based MRF algorithm that can detect road regions in  images and videos.
In our method, the date fidelity term of the energy function is combined with various types of superpixel features.
The smoothness terms are computed according to the spatio-temporal proximity of superpixels.\par

\textbf{(3) Scene model construction:}
3D modeling and reconstruction is currently a hot topic in computer vision and computer graphics communities.
With regard to  the reconstruction of single objects, a number of important achievements have now been realized in relation to 3D building reconstruction, 3D model retrieval, 3D face simulation, etc.
The general aim with all of these technologies is to recover the 3D geometric surface of single objects.
%Many state of the art methods are employed for these tasks: Shape from shading, shape from texture, structure from motion, etc.
As soon as one moves beyond  single object reconstruction, 3D reconstruction becomes much more challenging.
 %such as saliency computation and region detection.
%Region detection is the preliminary to the reconstruction of 3D scene models.
%In order to reconstruct the 3D scene models, region detection should be implemented firstly.
%The semantic labels of image regions are then specified to determine their locations and orientations in the 3D space.
%The corresponding scene models are constructed based on the semantic labels.
%Furthermore, new viewpoint images can be rendered based on the scene models.
%The TIP framework proposed by Horry et al.  \cite{Tour_Into_Picture} aims to recover the 3D structure from a single still image .
%After the detection of the vanishing point, the input image is roughly partitioned into ``left wall'', ``right wall'', ``back wall'' and ``road plane''.
%The foreground models can be constructed independently of the background models.
%However, the TIP model is not fit for the curved-edge road conditions.
Saxena et al. \cite{Saxena} have put forward a  plane parameters learning approach that is based on MRF.
This is used to judge the location and orientation of recovered mesh facets.
 Make3D models with wireframe meshes are then constructed.
The monocular image features used in this approach principally include color, texture, gradient and edges.
The plane and model correctness ratios are utilized as metrics to evaluate the scene models.
However, the method does not take into account the possible impact of large-scale semantic environments.
%As a result, the connection properties among superpixels must be judged.
Hoiem et al. \cite{Hoiem} have developed an approach that first  performs superpixel segmentation of input images, then applies support vector machine (SVM) to cluster superpixels with a similar appearance into cliques.
The regional properties of the superpixel cliques are then specified to construct  ``pop up'' style 3D scene models.
Lou et al. \cite{Scene_Stage} were the first to use layout templates to predict a global image structure.
A random walk algorithm is then used to infer refined scene structures.
However, the limited number of scene stages prevents this from  being applied to all road scene structures.

The rest of the paper is organized as follows:
In Section 3, an MRF model at pixel level is introduced.
Our approach to region detection using superpixel-based MRF is described in Section 4.
%The region detection method using superpixel MRF is described in Section IV.
Section 5 details the spatio-temporal scene reconstruction process.
The results of an experimental validation of the proposed framework are presented and discussed in Section 6.
%Experiments and discussions are conducted in Section VI.
Section 7 provides our conclusions and indicates our future works.

\section{MRF Model at Pixel Level}

We are first of all going to introduce the pixel-level MRF model and how it works.
% we introduce the MRF model at pixel level.
Two groups of random variables, $X$ and $Y$, are defined.
These refer to the observational data and state variables, respectively.
The posterior probability distribution, $P(Y|X)$, can be solved by using Bayesian theory \cite{Li_Stan}.
%\begin{equation}
%P(Y|X) = \frac{{P(X|Y)P(Y)}}{{P(X)}},
%\end{equation}
%where  $P(Y)$ is the prior probability of $Y$.
%$P(X|Y)$ is the conditional probability distribution given $Y$.
%$P(X)$ is the density function of $X$, which is a constant when $X$ is identified.\par
For the task of image annotation, $X$ denotes the pixel coordinate set in the image, while $Y$ denotes the corresponding set of labels.
The constraint between these variables is established to compute $P(Y)$, which is assumed to follow a uniform distribution.
%\begin{eqnarray}
%P(Y)=\frac{1}{Z}\exp(-(\sum_{i,j \in N_i}f_{ij}(y_i,y_j))),
%\label{equ_2}
%\end{eqnarray}
%where $f_{ij}(\bullet)$ denotes the uniform Gibbs distribution.
%The neighborhood domain of $i$ is denoted by $N_i$.
%$Z$ is used for normalization.\par
Assuming the elements in $X$ and $Y$ are independent of each other, the likelihood function $P(X|Y)$ can be defined as follows:
\begin{eqnarray}
P(X|Y)=\prod_i f_i(x_i,y_i)  =\exp(\sum_{i}\log f_i(x_i,y_i)),
\label{equ_3}
\end{eqnarray}
where $f_i(\bullet)$ is the data fidelity term of the $i_{th}$ element.\par

\begin{figure}[!htbp]
\centering
\includegraphics[width=7cm]{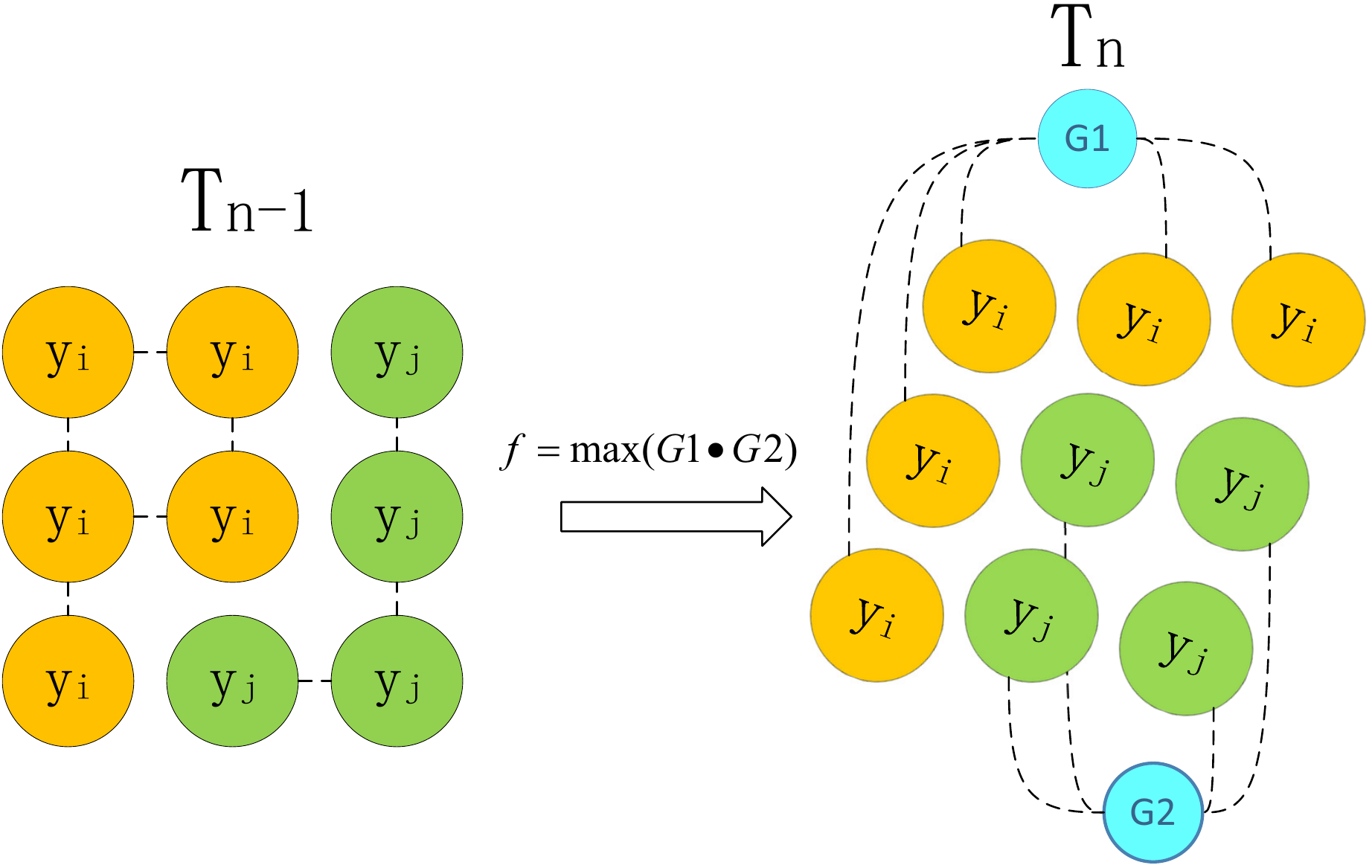}
\caption{Illustration of Gibbs distribution. }
\label{fig:Gibbs}
\end{figure}

Based on the above analysis, Eq.(2) can be defined by:
\begin{equation}
%\begin{split}
P(X|Y)P(Y)=\frac{1}{Z}\exp(\sum_{i}\log f_i(x_i,y_i))
\exp(-(\sum_{i,j \in N_i}f_{ij}(y_i,y_{j}))).
%\end{split}
\label{equ_4}
\end{equation}
where $f_{ij}(\bullet)$ denotes the uniform Gibbs distribution.
The labels of $y_i$ and $y_j$ can be transformed to each other in the Gibbs distribution, as shown in Fig. \ref{fig:Gibbs}.
The function of $f_{ij}$ makes the product of $G1$ (pixels of label $y_i$) and $G2$ (pixels of label $y_j$) to reach the maximum.\par

When the constant component is removed, the energy function is defined as follows:
\begin{eqnarray}
E(Y|X)=\sum_{i,j \in N_i}f_{ij}(y_i,y_j)-\sum_{i}\log f_i(x_i,y_i).
\label{equ_5}
\end{eqnarray}

The variable $Y$ is solved by the minimization of energy function:
\begin{eqnarray}
{Y^*} = \arg {\kern 1pt} {\kern 1pt} {\kern 1pt} {\kern 1pt} \mathop {\min }\limits_Y {\kern 1pt} {\kern 1pt} {\kern 1pt} E(Y|X).
\label{equ_6}
\end{eqnarray}

%The general solution process of the MRF method is as follows:\par
%\begin{itemize}
%\item Define the observation variable set $X$ and the initial state variable set $Y$;\par
%\item Specify the data fidelity term $f_i(\bullet)$ and the data smoothness term $f_{ij}(\bullet)$;\par
%\item Solve the state variable $Y^*$ based on the energy minimization algorithm.
%\end{itemize}

Several classic algorithms can be used for energy minimization, such as greedy algorithms and dynamic programming algorithms.
A pixel-level MRF is useful for the image segmentation tasks, saliency detection, \emph{etc}.
%The pixel-based MRF algorithm has important applications in the community of computer vision.
%Image segmentation, image saliency detection are typical examples of such applications.
%However, the pixel-level MRF model has two main disadvantages: (1) too much time cost and (2) low dimension of vision feature vectors.
%High time-consuming and low dimensions of vision feature.
However, pixels that are captured by a camera sensor are not meaningful units by themselves.
They are also very susceptible to noise \cite{Schick_A}.
In order to improve their representational efficiency, we decided to use a new MRF model based on superpixels.
The main merit of superpixels is that they provide a more natural and perceptually meaningful representation of an input image \cite{Jianbing_Shen}. 
A superpixel representation greatly reduces the number of image primitives and is more robust to noise. 
Computation of a region's visual features by superpixels is also more effective.\par
%Moreover, it is more effective to compute the region based visual features by superpixels.\par

\section{Spatio-temporal road region detection}

Having established the pixel-level MRF model, we will now present the new superpixel-based MRF model,
beginning with superpixel segmentation and feature pool construction.
%The superpixel segmentation and feature pool construction are first implemented for the proposed method.
%For the superpixel-level MRF modeling, we first perform superpixel segmentation for each input image.
%According to the different definition types  of the smoothness term, the MRF model can be classified into two groups:
%(1)Superpixel MRF model for single images, where the smoothness term is computed in spatial domain;
%(2)Superpixel MRF model for image sequences, where the smoothness term is computed in both spatial and temporal domains.
%For the first definition type, the algorithm is implemented independently for each frame.
%After the definition of the energy function, the energy minimization is implemented.
%For the second definition type, the superpixel interactions in temporal domain are also incorporated for the smoothness term.
% definition of the smoothness term

\subsection{Superpixels and Feature Pools}

%Firstly, we will introduce the feature descriptions of the superpixels.
%The feature pools based on superpixels are subsequently constructed.
%Furthermore, the superpixel-based MRF algorithms for single image and image sequences are proposed respectively.

%\subsubsection{Descriptions of Superpixel Features}

Thousands of superpixels may exist in a single still image.
These are taken as mid-level image units.
Useful image features can be generated by clustering the superpixels into cliques.
Ideally, a superpixel clique will occupy an image region that represents a certain semantic object.
However, the clique may not correspond to a semantic object even if all the superpixels have the same label.
Superpixel features are widely used to assist with the tasks of image segmentation, semantic annotation, \emph{etc.}
These features primarily include color, texture and geometric shape, as shown in  Table \ref{tab:superpixel_feature}.
 %shows the summarization of these features.

\begin{table}[!hbtp]
\centering
\caption{Feature Descriptions of Superpixels and Superpixel Cliques}
\begin{tabular}{|l|c|}
\hline
\textbf{Feature Descriptors} & \textbf{Feature Numbers} \\
\hline
\textbf{Color}    &\textbf{9}\\
C1 RGB color      &3\\
C2 HSV color      &3\\
C3 CIELAB         &3\\
\hline
\textbf{Texture}  &\textbf{62}\\
T1 Gabor filters: 4 scales, 6 orientations  &48\\
T2 Local binary pattern: $3\times3$ template  &9\\
T3 Edge histogram descriptors               &5\\
\hline
\textbf{Locations and Shapes}  &\textbf{6}\\
L1 Location: Normalized $x$ and $y$ coordinates  &2\\
L2 Shapes: Superpixel number in the clique     &1\\
L3 Shapes: Edge number within convex hull            &1\\
L4 Shapes: Ratio of the pixels to the convex hull &1\\
L5 Shapes: Whether the clique is continuous       &1\\
\hline
\end{tabular}
\label{tab:superpixel_feature}
\end{table}

\begin{figure}[t]
\centering
\includegraphics[width=10cm]{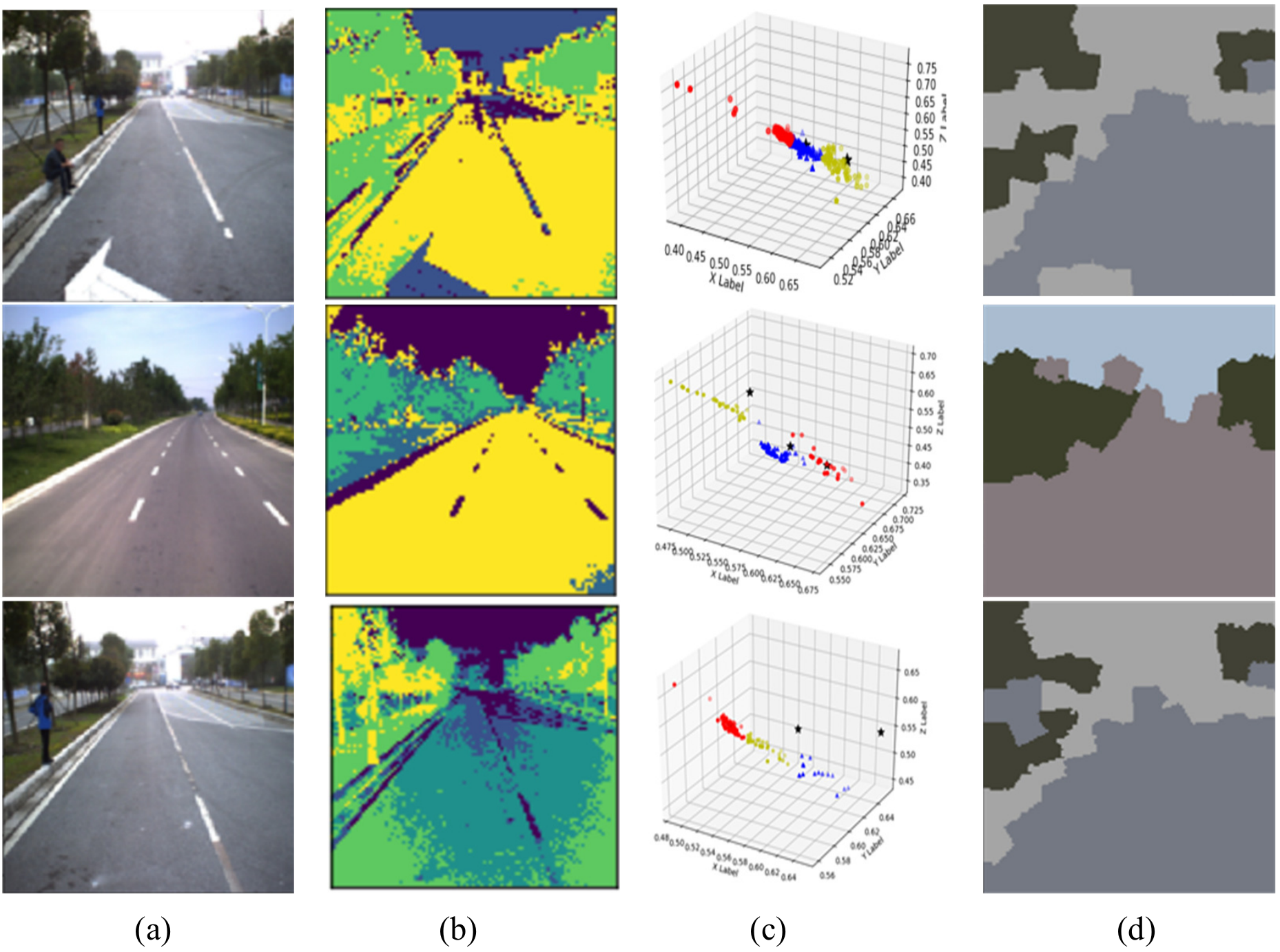}
\caption{Construction of color feature pools. (a) Input images. (b) Feature clustering results using K-means. (c) 3D representation of feature clustering using SOM. (d) Feature clustering results using SOM. }
\label{fig:SOM_cluster}
\end{figure}

Color feature pools can be constructed using a self-organizing map (SOM) based on competitive learning \cite{SOM}.
The main advantage of using an SOM method rather than K-means clustering is the adaptive clustering it provides.
Color feature clustering based on SOM involves three processes, competition, cooperation and adaptation, in which the Euclidean distance between input and output data is utilized.
The mean and standard error for each superpixel belonging to a certain image region are computed in the CIELab space.
%For each superpixel belonging to a certain image region, its mean and standard error in the CIELab space are computed as features.
The input is the superpixel color information within road regions in a set of training images.
% to the SOM are the superpixel features in the training images.
After specification of the neurons in the SOM  network, the clustering means are computed without  preassigning the clustering mean  number. 
The SOM clustering results are assumed to be the color means in the feature pool.
Fig. \ref{fig:SOM_cluster} shows examples of color feature clustering using SOM.\par

To represent the texture features, the local texture is represented by the intensity distribution of adjacent pixels.
The global texture is defined by  assembling the local texture information.
The texture feature pools are constructed by clustering the Gabor filter outputs for all of the image regions \cite{Log_Gabor}.
For each pixel in the image, a corresponding Gabor filter output vector is computed.
%These output vectors preserve the energy and ignore the location information.
We chose to implement a Gabor filter with 4 scales and 6 directions.
So, the output vector dimension for a $128\times128$ sized Gabor filter is $128\times128\times4\times6=393216$.
The computation for such high-dimensional feature vector is very time-consuming.
Besides, serious feature redundance can be suffered.
So, we only chose to concentrate on just the mean and the standard error of the Gabor filter output.
As a result, there is only $2\times4\times6=48$ feature dimension to be concerned with.
%the feature dimension is only $2\times4\times6=48$ under this case.
%The texture feature pools are generated by K-means clustering of texture features in different image regions.

\subsection{Road Region Detection}

\begin{figure}[t]
\centering
\includegraphics[width=13cm]{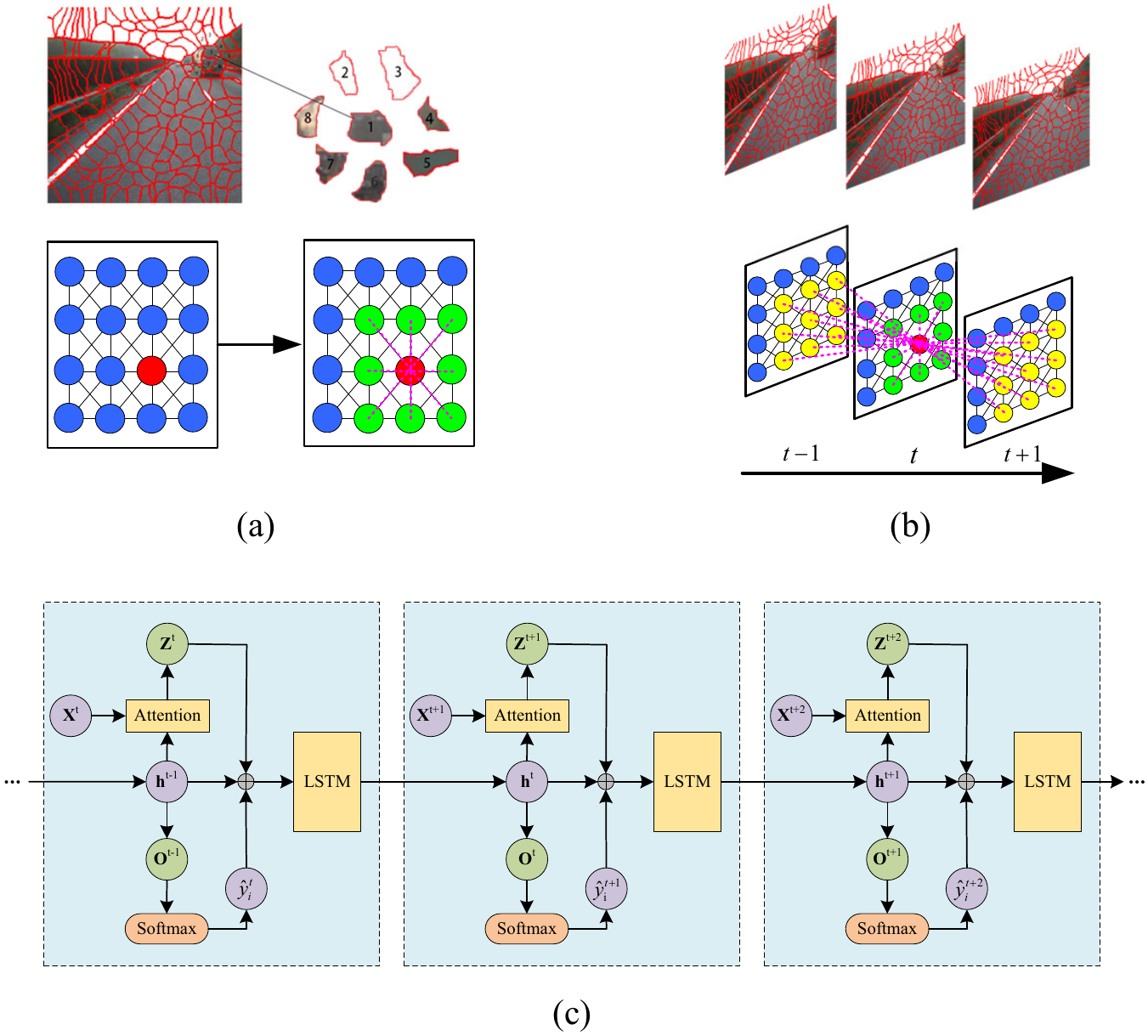}
\caption{Superpixels and label initialization. (a) Superpixel neighborhoods in spatial domain. (b) Superpixel neighborhoods in spatio-temporal domain. (c)Supperpixel label initialization using LSTM.}
\label{fig:superpixel_temporal}
\end{figure}

\begin{figure*}[!htbp]
\centering
\includegraphics[width=14cm]{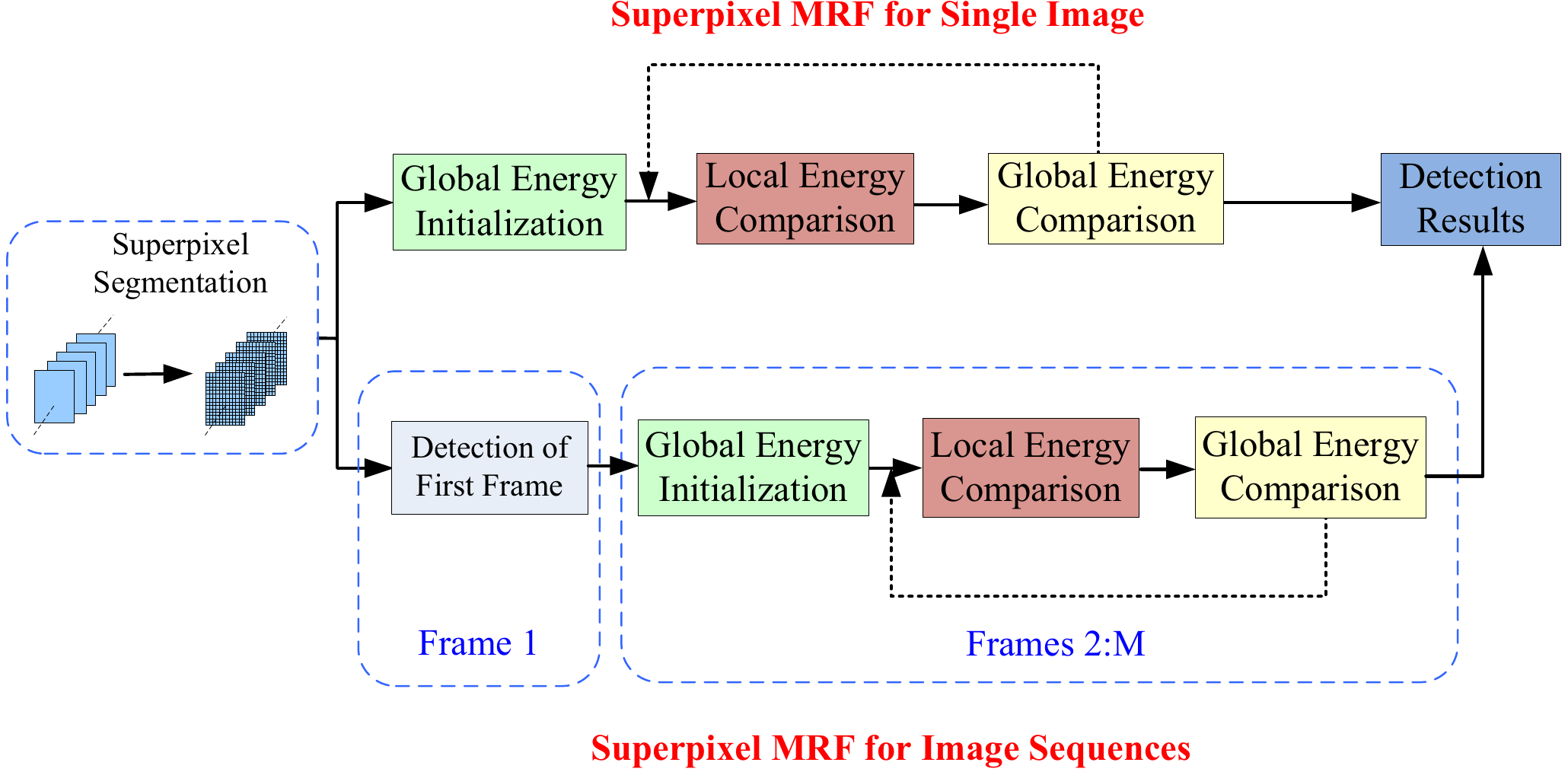}
\caption{Flow diagram of the road detection process.}
\label{fig:road_detection}
\end{figure*}

After constructing the feature pools, we introduce a road region detection algorithm that utilized the superpixel-based MRF.
%superpixel MRF method for road region detection.
The connections between spatially and spatio-temporally adjacent superpixels are shown in Fig. \ref{fig:superpixel_temporal} (a) and (b).
The red dots in the figure denote the current superpixel for processing.
The green dots denote the spatially adjacent superpixels.
The yellow dots relate to adjacent superpixels in both spatial and temporal domains.

Fig. \ref{fig:road_detection} shows the flow diagram of the road detection process, which works for both single images and image sequences.
%Firstly, we propose the road detection algorithm for single images.
We will focus here on the algorithm for image sequences.
In this algorithm, the global energy for the superpixel-based MRF is as follows:
\begin{eqnarray}
%\begin{split}
E_1^t= -\sum_{i\in S}f_i(\bold{x}_i^t|y_i^t)+\sum_{i\in S}\sum_{j \in N_{spa}\{i\}}\lambda_1f_{ij}(y_i^t,y_j^t)
      +\sum_{i\in S}\sum_{j \in N_{tem}\{i\}}\lambda_2f_{ij}(y_i^t,y_j^t)
%\end{split}
\label{equ_17}
\end{eqnarray}
where $\mathcal{S}$ denotes the superpixel set;
$\bold{x}_i^t$ is the appearance feature of the $i_{th}$ superpixel at time $t$;
$y_i^t$ denotes the label of the $i_{th}$ superpixel, which is in the set of $\mathcal{L}=\{0,1\}$;
$\log p(\bold{x}_i^t|y_i^t)$ denotes the data fidelity term of the $i_{th}$ superpixel;
$f_{ij}(\bullet)$ is the smoothness term of the adjacent superpixel pairs;
$N_{spa}\{i\}$ and $N_{tem}\{i\}$ denote the adjacent neighbors of superpixel $i$ in spatial and temporal domains, respectively;
$\lambda_1$ and $\lambda_2$ are constants.\par

The probability of the data fidelity term is computed as follows, using a combination of color, texture and location information for any time $t$:
\begin{eqnarray}
p(\bold{x}_i|y_i)=p(\bold{c}_i|y_i)p(\bold{t}_i|y_i)p(\bold{lc}_i|y_i),
\label{equ_9}
\end{eqnarray}
where $\bold{c}_i$, $\bold{t}_i$ and $\bold{lc}_i$ denote the color, texture and location features of the $i_{th}$ superpixel, respectively.\par

The color probability is defined by the following three-channel Gaussian distribution:
\begin{eqnarray}
%\begin{split}
p(\bold{c}_i|y_i)=\mathop{sup}_{\{\boldsymbol{\mu}_m,\boldsymbol{\Sigma_{m}\}}\in C_r} \{\frac{1}{(2\pi)^{d/2}|\boldsymbol{\Sigma_{m}}|^{1/2}}
\exp (-\frac{1}{2}(\bold{c}_i-\boldsymbol{\mu}_m)^T\boldsymbol{\Sigma_{m}}^{-1}(\bold{c}_i-\boldsymbol{\mu}_m))\},
%\end{split}
\label{equ_10}
\end{eqnarray}
where $\boldsymbol{\mu}_m$ and $\boldsymbol{\Sigma_m}$ are the $m_{th}$ mean and covariance in the color feature pool corresponding to  label $y_i$.
%�ֱ�������ǩ$y_i$��Ӧ����ɫ�������У���$m$����ֵ��Э����������

The road region usually contains a rich variety of textural materials and stripes.
The texture features can be used to compute the probability of the data fidelity term as an additional measure.
We employ a Gabor filter to extract the texture features.
The image block centered upon each superpixel is used to compute the output vectors of the Gabor filter.
The computation is implemented by using a correlation coefficient together with the cluster means in the texture feature pool.
The exponential form of the largest coefficient is defined as the texture probability:
\begin{eqnarray}
p(\bold{t}_i|y_i)=\mathop {sup}_{\bold{MT}_m \in T_r}   \exp(r(\bold{t}_i,\bold{MT}_m)),
\label{equ_11}
\end{eqnarray}
where $\bold{MT}_m$ denotes the $m_{th}$ clustering mean in the texture feature pool $T_r$;
and $r(\bullet)$ is the correlation coefficient:
\begin{equation}
r({{\bf{t}}_i},{\bf{M}}{{\bf{T}}_m}) = \frac{{N\sum\limits_{k = 1}^N {{{\bf{t}}_i}[k] \cdot {\bf{M}}{{\bf{T}}_m}[k]}  - \sum\limits_{k = 1}^N {{{\bf{t}}_i}[k] \cdot \sum\limits_{k = 1}^N {{\bf{M}}{{\bf{T}}_m}[k]} } }}{{\sqrt {(N\sum\limits_{k = 1}^N {{\bf{t}}_i^2[k] - {{(\sum\limits_{k = 1}^N {{{\bf{t}}_i}[k]} )}^2}} ) \cdot (N\sum\limits_{k = 1}^N {{\bf{MT}}_m^2[k] - {{(\sum\limits_{k = 1}^N {{\bf{M}}{{\bf{T}}_m}[k]} )}^2}} )} }}
\end{equation}
where $N$ denotes the dimension of texture feature vector.\par

The location probability is computed as follows:
\begin{eqnarray}
p({{\bf{lc}}_i}|{y_i}) = \exp ({(\frac{{N{M_{{y_i},{{\bf{lc}}_i}}} + {\alpha _\lambda }}}{{N{M_{{{\bf{lc}}_i}}} + {\alpha _\lambda }}})^{{\omega _\lambda }}}).
\label{equ_12}
\end{eqnarray}\par

For the computation of the location probability, the input image is projected onto a rectangle.
$\bold{lc}_i$ denotes the locations in the rectangle.
For the training set, the same projection method is applied.
${N{M_{{y_i},{{\bold{lc}}_i}}}}$ denotes the number of superpixels belonging to the label $y_i$,  corresponding to the location $\bf{lc}_i$ in the regular rectangle.
${N{M_{{{\bf{lc}}_i}}}}$ denotes the total number of superpixels in  location $\bold{lc}_i$ of the regular rectangle.
$\alpha _\lambda$ is set to be a certain small value, \textit{e.g.} $\alpha _\lambda=0.5$.
The  constant, $\omega _\lambda$, varies according to the different datasets. \par

After computation of the data fidelity term, the smoothness term can be defined as:
\begin{equation}
{f_{ij}}({y_i},{y_j}) = (1 - \delta ({y_i},{y_j}))\exp ( - \beta  \cdot {d_M}(i,j))
\end{equation}
where ${d_M}(i,j)$ denotes the distance between the superpixel feature vectors:
\begin{equation}
{d_M}(i,j) = {({{\bold{x}}_i} - {{\bold{x}}_j})^T}\bold{M}({{\bold{x}}_i} - {{\bold{x}}_j})
\end{equation}
where $\bold{x_i}$ and $\bold{x_j}$ are the color features of $i_{th}$ and $j_{th}$ superpixels;
%$\| \bullet \|$ denotes the $L_2$ norm.
$\bold{M}$ is a positive semi-definite matrix that parameterizes the metric.
The metric is  Euclidean norm when $\bold{M}=\bold{I}$.
The metric is the Mahalanobis distance when $\bold{M} = {\bold{\Sigma} ^{ - 1}}$, where $\bold{\Sigma} ^{ - 1}$ denotes the inverse covariance matrix.
$\delta(\bullet)$ is the Kronecker delta:
\begin{eqnarray}
\delta ({y_i},{y_j}) = \left\{ \begin{array}{l}
 0{\kern 1pt} {\kern 1pt} {\kern 1pt} {\kern 1pt} {\kern 1pt} {\kern 1pt} {\kern 1pt} \mbox{if}{\kern 1pt} {\kern 1pt} {\kern 1pt} {\kern 1pt} {y_i} \ne {y_j} \\
 1{\kern 1pt} {\kern 1pt} {\kern 1pt} {\kern 1pt} {\kern 1pt} {\kern 1pt} {\kern 1pt} \mbox{if}{\kern 1pt} {\kern 1pt} {\kern 1pt} {\kern 1pt} {y_i} = {y_j} .\\
 \end{array} \right.
 \label{equ_14}
\end{eqnarray}

The constant coefficient, $\beta$, can be defined as follows:
\begin{eqnarray}
\beta=(2<\|\bold{x_i}-\bold{x_j}\|^2>)^{-1},
\label{equ_15}
\end{eqnarray}
where $<\bullet>$ denotes the expectation of the superpixel pairs.\par

In addition to the global energy function, the local energy function for the $i_{th}$ superpixel is as follows:
\begin{eqnarray}
%\begin{split}
E_{1i}^t =  - {f_i}({{\bf{x}}^t_i}|{y^t_i}) + \sum\limits_{j \in {N_{spa}}\{ i\} } {{\lambda _1}{f_{ij}}({y^t_i},{y^t_j}) }
 + \sum\limits_{j \in {N_{tem}}\{ i\} } {{\lambda _2}{f_{ij}}({y^t_i},{y_j^t})}.
%\end{split}
\label{equ_18}
\end{eqnarray}

\begin{figure}[t]
\centering
\includegraphics[width=11.5cm]{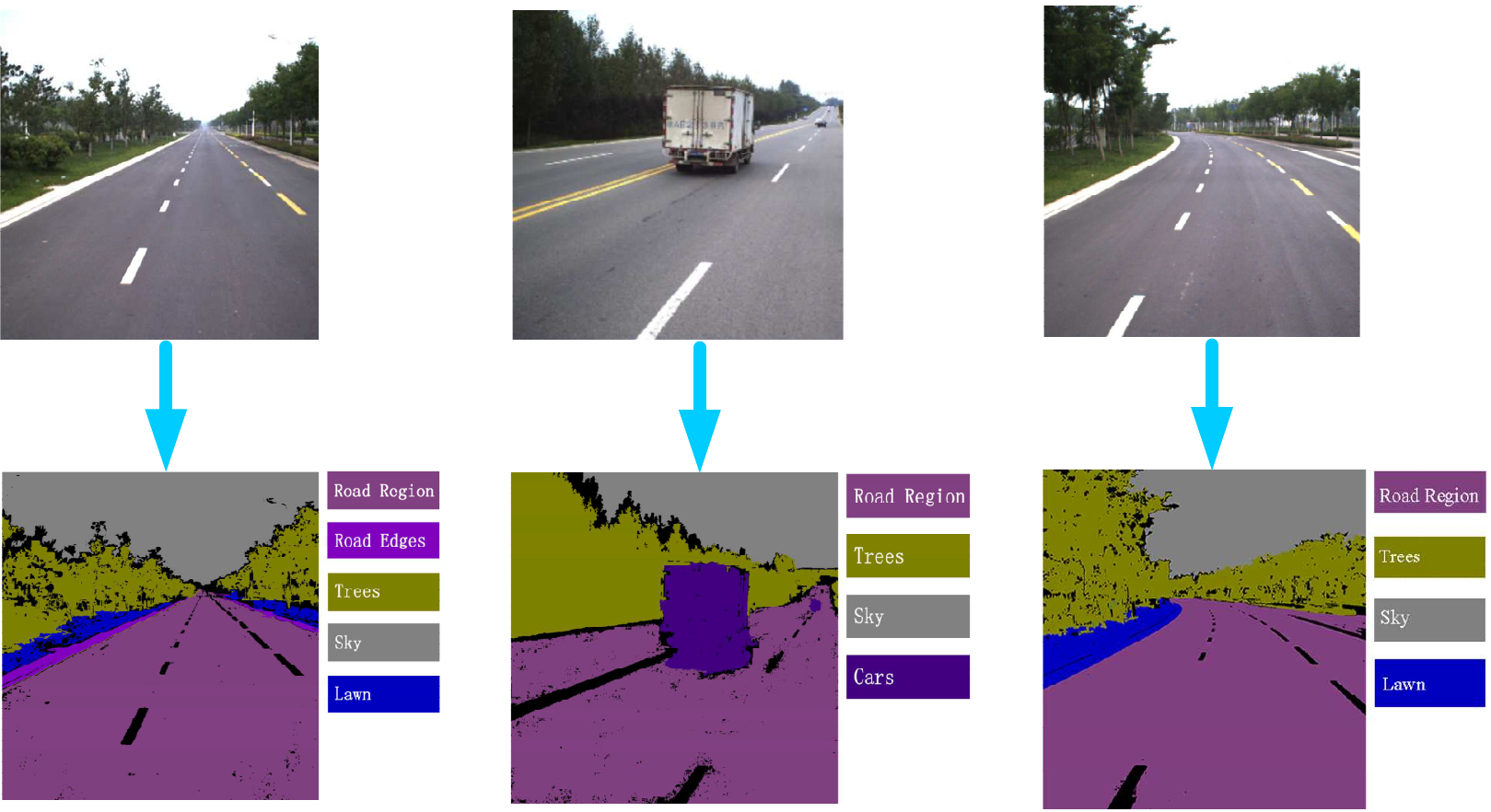}
\caption{Superpixel label initialization. }
\label{fig:semantic_labeling}
\end{figure}

Working on the assumption that $M$ frames exist in the image sequence, the proposed superpixel-based MRF algorithm is shown in Algorithm 1.
The label initialization for the first frame is implemented using a semantic labeling method \cite{Semantic_Object_Classes}, as shown in Fig. \ref{fig:semantic_labeling}.
The semantic labeling method incorporates three different image segmentation methods: GraphCut \cite{Graph_based_Segment}, MeanShift \cite{Mean_Shift} and Image Pyramid method \cite{Segment_and_Estimation}.
%By the scheme of competition and cooperation, the three segmentation methods obtain an optimized labeling result.
%The three segmentation methods obtain the optimized semantic labeling result by the scheme of competition and coordination.
Although this semantic labeling method achieves a more accurate initialization, we only chose to apply it to initialize the first frame, because it requires too much human interaction and is very time-consuming.\par

For the rest of the frames, the superpixel initialization is implemented by a long short-term memory (LSTM) network, as shown in Fig. \ref{fig:superpixel_temporal} (b).
Beginning with the LSTM network training process, for the $i_{th}$ superpixel in the $t_{th}$ frame, the LSTM input data includes the superpixel feature vector $\bold{X}^t$, and the corresponding ground truth superpixel label $y_i^t$, where $\bold{X}^t=[\bold{x}_1^t, \bold{x}_2^t,...,\bold{x}_K^t]$; $K$ denotes the superpixel number.
The output of the LSTM network $\hat{y}^{t+1}_{i}$ is compared with the ground truth superpixel label $y^{t+1}_i$,
and a cross entropy function is used to compute the error.\par

After this, the trained LSTM network is used for the initialization of the superpixel labels for each frame.
The input to the LSTM is again $\bold{X}^t$ for the $i_{th}$ superpixel in frame $t$.
$\bold{h}^{t-1}$ and $\bold{h}^t$ denote the hidden states of the LSTM network at time $t-1$ and $t$, respectively.
$\bold{Z}^t$ is the weighted feature based on the attention mechanism \cite{Attention}.
$\bold{O}^t$ is the output of the LSTM network at time $t$.
$\hat{y}^t_i$ and $\hat{y}^{t+1}_{i}$ are the initialized labels for the $i_{th}$ superpixel at times $t$ and $t+1$, respectively.\par

%The initial superpixel labels are specified according to the prior knowledge, \textit{e.g.} the bottom $1/3$ part of the image can be initialized to be the road region.
%Another option is to initialize the center part of the image as road region, while the non-road region is defined by:
%\begin{eqnarray}
%B =  \{ \mathcal{S}(i)|\min (m,n,|W - m|,|H - n|) \le \omega ,{\kern 1pt} {\kern 1pt} {\kern 1pt}
%     (m,n) \in \mathcal{S}(i)\}.
%\label{equ_16}
%\end{eqnarray}
%where $W$ and $H$ denote the width and height of the image.
%$\omega$ is the width of the surrounding region.
%$\mathcal{S}(i)$ denotes the $i_{th}$ superpixel.
%$(m,n)$ denotes the pixel coordinates within the superpixel.

\begin{algorithm}[t]
%\renewcommand{\captionfont}{\wuhaoalg \wuhaoalg \wuhaoalg}
%\vspace{1ex}
\caption{Superpixel MRF for image sequence}
\begin{algorithmic}[1]
\REQUIRE $\begin{cases}
\text{Input image sequence $\{I_1,...,I_M\}$;}\\
\text{Superpixel set for each frame $\{\mathcal{S}_1,...,\mathcal{S}_M\}$;}\\
\text{Threshold of the global energy function $\varepsilon$.}
\end{cases}$
\STATE    \textnormal{Specify the initial superpixel label set of first frame;}
\STATE    \textnormal{Compute the initial global energy $E_1^1$ based on the initial superpixel labels;}
\FOR {$t=2:M$}
\STATE    \textnormal{Initialize the superpixel label set of time $t$ using the LSTM network;}\\
\STATE    \textnormal{Compute the initial global energy function $E_1^t$ of current frame according to the initial superpixel class labels. The energy function is composed of data fidelity term and smoothness term; }\\
\STATE    \textnormal{For each superpixel $i$, compute its local energy $E_{1i}^t$;
The $\alpha-\beta$ swap is performed to compute the new local energy $\hat{E_{1i}^t}$;
If  $\hat{E_{1i}^t}<E_{1i}^t$, update the label: $y_i=\mathcal L \backslash y_i$};\\
\STATE    \textnormal{Compute the updated global energy $\hat{E_1^t}$ according to the new class labels};\\
\STATE    \textnormal{If $\hat{E_1^t}$ and $E_1^t$ are within the threshold distance of $\varepsilon$, the algorithm terminates; Otherwise jump to Step 6;}\\
\ENDFOR
\STATE    \textnormal{Algorithm stops.}

\ENSURE Superpixel label set for all the image frames $\{\bold{Y}_1,...\bold{Y}_M\}$.
\end{algorithmic}
\end{algorithm}

After the superpixel initialization step, the road detection algorithm is designed to follow a cycle of ``initial energy computation-local energy comparison-global energy comparison''.
The smoothness term of the energy function incorporates the superpixel interactions in both the spatial and temporal domains.
%As a result, the computation of local energy is more accurate.
We assume that the road region between adjacent frames has barely changed, and apply the  superpixel centers from the previous frame to initialize the current one.
An optical flow map is also used to match the superpixels in adjacent frames, in case the road region between adjacent frames has changed more significantly.
The optical flow map is computed according to Bruhn's method \cite{Bruhn}, which is applied to align the superpixel centers of the current frame with  the nearest centers in adjacent frames.\par

%Our superpixel MRF algorithm has several steps, e.g. local and global energy comparisons.
% so the initialization of superpixel labels is of no great importance to the algorithm.\par

Drawing upon the initial superpixel labels, the initial global energy function $E_1$ can be computed using Eq. (7).
The value of $E_1$ is the sum of the local energy function $E_{1i}$.
The energy function is composed of a data fidelity term and  a smoothness term.
The data fidelity term is defined by a combination of color, texture and location probabilities.
The smoothness term is defined by the interaction between spatially adjacent superpixels (Eq. 13).
After computation of the global energy function, the local energy $E_i^t$ of superpixel $i$ is calculated according to Eq. (8).
An $\alpha-\beta$ swap is then applied for the label $y_i$, and  the new local energy $\hat{E_i}^t$ is compared to $E_i^t$.
%We then apply $\alpha-\beta$ swap for the label $y_i$, and compare the new local energy $\hat{E_i}$ with $E_i$.
If the new local energy has reduced, the current class label is displaced.
The $\alpha-\beta$ swap method has a similar mechanism as $\alpha$-expansion \cite{alpha_beta}, which is widely used in energy minimization.
After  the comparison of  all the local energies, the  global energy $\hat{E_1^t}$ is updated.
If  $\hat{E_1}^t$ and the previous value of $E_1^t$ are within a small threshold distance $\varepsilon$, the algorithm terminates.

%\subsection{Road Detection for Image Sequences}
%As mentioned above, the superpixel MRF algorithm for single images has its drawbacks when dealing with image sequences.
%For further improvement, we incorporate the spatial-temporal coherent information of image sequences.

%\STATE    \textnormal{���ڳ�ʼ������������ǩ��������ʼȫ����������$E_1$�������������������ݹ۲��������ݽ��������������ɣ��ֱ����Զ����ͼ��㣻}\\
%\STATE    \textnormal{����ÿ��������$i$�����ֲ�����ֵ$E_{1i}$��������ǩֵ$y_i$ ����$\alpha-\beta$�û������������õľֲ�����ֵ$\hat{E_{1i}}<E_{1i}$������������ǩ���ֲ�����ֵ:$y_i=\mathcal L \backslash y_i$��${E_{1i}}=\hat{E_{1i}}$������������ǰ��ǩ���ֲ�����ֵ��}\\
%ͨ�����µ�ǰ��ǩ�����𣬼����µľֲ�����ֵ������ԭ�оֲ�����ֵ���бȽϣ������µľֲ�����ֵ��С������ǰ������ǩ�����滻������������ǰ������ǩ��}\\
%\STATE    \textnormal{�����µ�������ǩ���ֲ�����ֵ���������º���ȫ������ֵ$\hat{E_1}$��}\\
%\STATE    \textnormal{����$\hat{E_1}$���ϴε���ȡֵ$E_1$ ֮��С����ֵ$\varepsilon$�����㷨��ֹ��������ת������3ѭ��ִ��;}\\
%\STATE    \textnormal{�㷨��ֹ.}

The proposed algorithm can be implemented for each single image independently.
Under such conditions, the smoothness term of the energy function only takes into account the spatially neighboring superpixels.
However, running the algorithm for complete image sequences of images offers certain advantages:
\begin{itemize}
\item Smoothness term.
      %In the superpixel MRF method for single images, the smoothness term only deals with the neighborhood superpixels within current frame.
      %The continuity property of superpixel features in temporal domain is neglected.
      The smoothness term for image sequences incorporates the neighboring superpixels in both the spatial and temporal domains.

\item Initialization of the labels for the first frame.
      The superpixel-based MRF method for image sequences applies more accurate semantic labeling to initialize the first frame.
      %This lays a solid foundation for the propagation of superpixel labels between adjacent frames.

\item Initialization of the labels for the rest of the frames.
      In the superpixel-based MRF method for image sequences, the LSTM network is applied to initialize the superpixel labels for the rest of the frames.
% superpixel labels of previous frame is propagated to current frame for initialization.
      %, the road region of adjacent frames is assumed to be almost unchanged.
     % As a result, we propagate the superpixel labels of previous frame to current frame as initialization.
      %This process improves the accuracy of label initialization.
      %Furthermore, the iteration number of the algorithm decreases and thus speed up the algorithm.

\end{itemize}

\section{Spatio-Temporal Scene Reconstruction}
%The road regions of the input image sequences are detected using the proposed superpixel MRF algorithm.
%Based on the detected road regions, the 3D corridor-structured scene models are constructed.
%The construction of scene models is based on the control points of road boundaries.
%\subsection{Control Points Generation}
After detecting the road regions in the input image sequences, control points on the road boundaries are generated to represent the road geometry.
%After the detection of the road regions in the input image sequences, the control points on road boundaries are generated to represent the road geometry.
Then, the road scene models are constructed based on the control points.
Several steps are needed to generate these control points.
Firstly, the vanishing point for each input image is detected by a Gaussian sphere method \cite{Bernard_ST}.
%The vanishing points are the intersections of the parallel lines in world coordinate system projected onto the image plane.
%With the assumption that the road boundaries are parallel in world coordinate system, the intersection point of the projected lines on image plane is considered to be the vanishing point.
%The detection of vanishing points mainly include three methods: (1) edge  based, (2) region  based and (3) texture based.
%The  Hough transform is applied via Gaussian sphere to detect the vanishing points\cite{Bernard_ST},
%which decompose the task of vanishing point detection into two steps: (1) line detection and (2) line pair detection.
%The image plane is projected onto a unit Gaussian sphere centered at the camera origin.
%The Canny operator is applied to line detection.
%The intersection points of parallel lines are corresponded to the curve intersections on the Gaussian sphere.\par
Secondly, the detected road regions are then processed for image binarization.
After this, the road boundaries can be generated based on the road and non-road regions.
%After the binarization of road and non-road regions, the road boundary curves are generated.
%We use a differential operator to detect the pixels with step change, and the pixels on the image edges are removed from the contour curves.
%The curvatures for the remaining pixels are computed, and the pixels with top curvatures are selected as candidate control points in maximum distance.
Given the constraint of  thevanishing points, the potential farthest control points can be identified. 
The points where the road boundaries intersect with the image edges are defined as the nearest control points.
%areas for the farthest control points are identified.
%The intersections of road boundaries with image edges are defined as the nearest control points.
The other control points are assumed to be uniformly distributed between the nearest and farthest control points.
%The examples of vanishing points and control points generation are shown in Fig. \ref{fig:control_points}.\par
%\begin{figure}[!htbp]
%\centering
%\includegraphics[width=8.5cm]{figures/control_points_generation_4.eps}
%\caption{Generation of vanishing points and control points. The yellow dots denote the vanishing point, while the red dots denote the control points of road boundaries.}
%\label{fig:control_points}
%\end{figure}
%\subsection{Scene Reconstruction}
A 3D corridor-style scene model can then be constructed based on the control points.
%After we generate the control points of road boundaries, the corridor-style scene models can be constructed.
%\subsubsection{Scene Reconstruction for Monocular Images}
We apply the Manhattan world assumption \cite{Manhattan} to specify the world coordinate system.
The scene models are assumed to follow a  Cartesian coordinate system.
Viewers can get  interpretations of the road scene by changing the alignment of the system.\par

%The three unit vectors $\bold{x}$, $\bold{y}$ and $\bold{z}$  are applied to define the camera axes, according to the Manhattan world system.
%The horizontal and vertical directions are defined by $\bold{x}$ and $\bold{y}$.
%$\bold{z}$ is the unit vector that satisfies  $\bold{x}\times\bold{y}=\bold{z}$.\par

\begin{algorithm}[t]
%\renewcommand{\captionfont}{\wuhaoalg \wuhaoalg \wuhaoalg}
%\vspace{1ex}
\caption{Road scene models construction}
\begin{algorithmic}[1]
\REQUIRE
Superpixel label sets for the image sequence:$\{\bold{Y}_1,...\bold{Y}_M\}$.
\FOR {$t=1:M$}
\STATE    \textnormal{Load the superpixel label set at frame $t$ to specify the road and non-road regions;}\\
\STATE    \textnormal{Detect the vanishing point using the Gaussian sphere method; }\\
\STATE    \textnormal{Generate the control points of road boundaries;}\\
\STATE    \textnormal{Perform  ``foreground segmentation/background inpainting'';}\\
\STATE    \textnormal{Background scene model is constructed according to the control points;}\\
\STATE    \textnormal{Foreground scene model is constructed with polygons of $RGBA$ data structure;}
\ENDFOR

\ENSURE Road scene models with the 3D Corridor structure.
\end{algorithmic}
\label{Alg_Scene}
\end{algorithm}

\begin{figure}[!htbp]
\centering
\includegraphics[width=14cm]{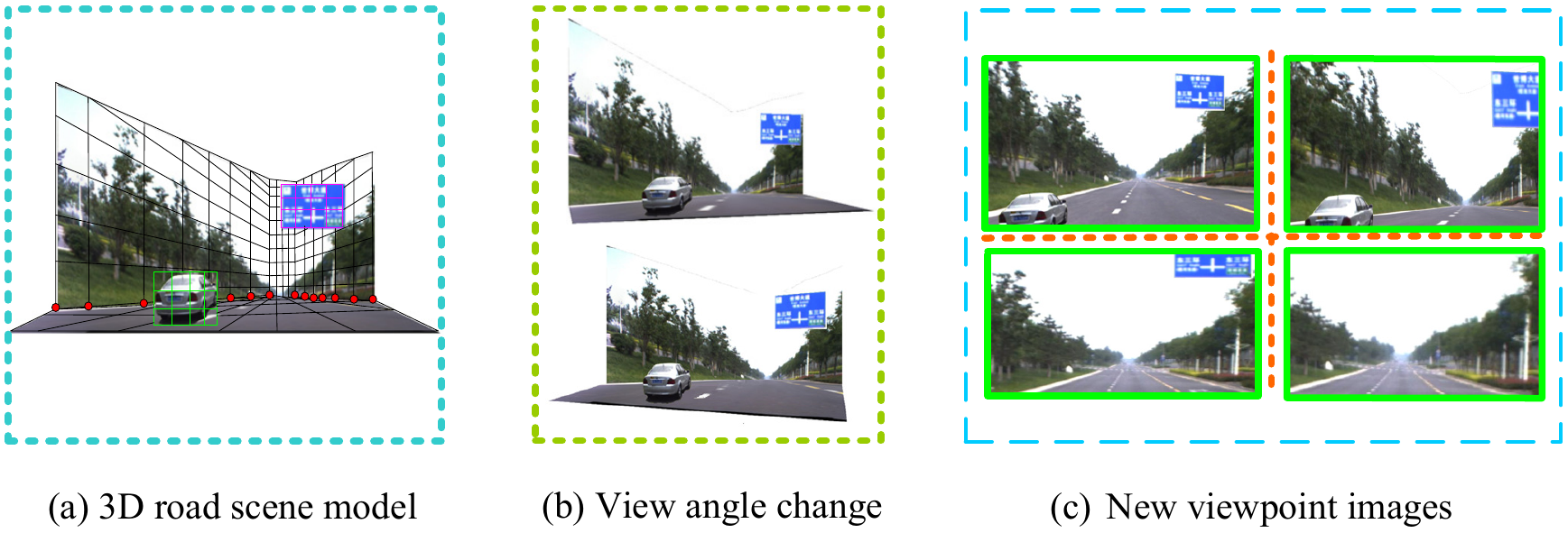}
\caption{Generation of new viewpoint images. (a) 3D road scene model.(b) The control of view angles. (c) The new viewpoint images. }
\label{fig:new_viewpoint}
\end{figure}

\begin{figure}[t]
\centering
\includegraphics[width=12cm]{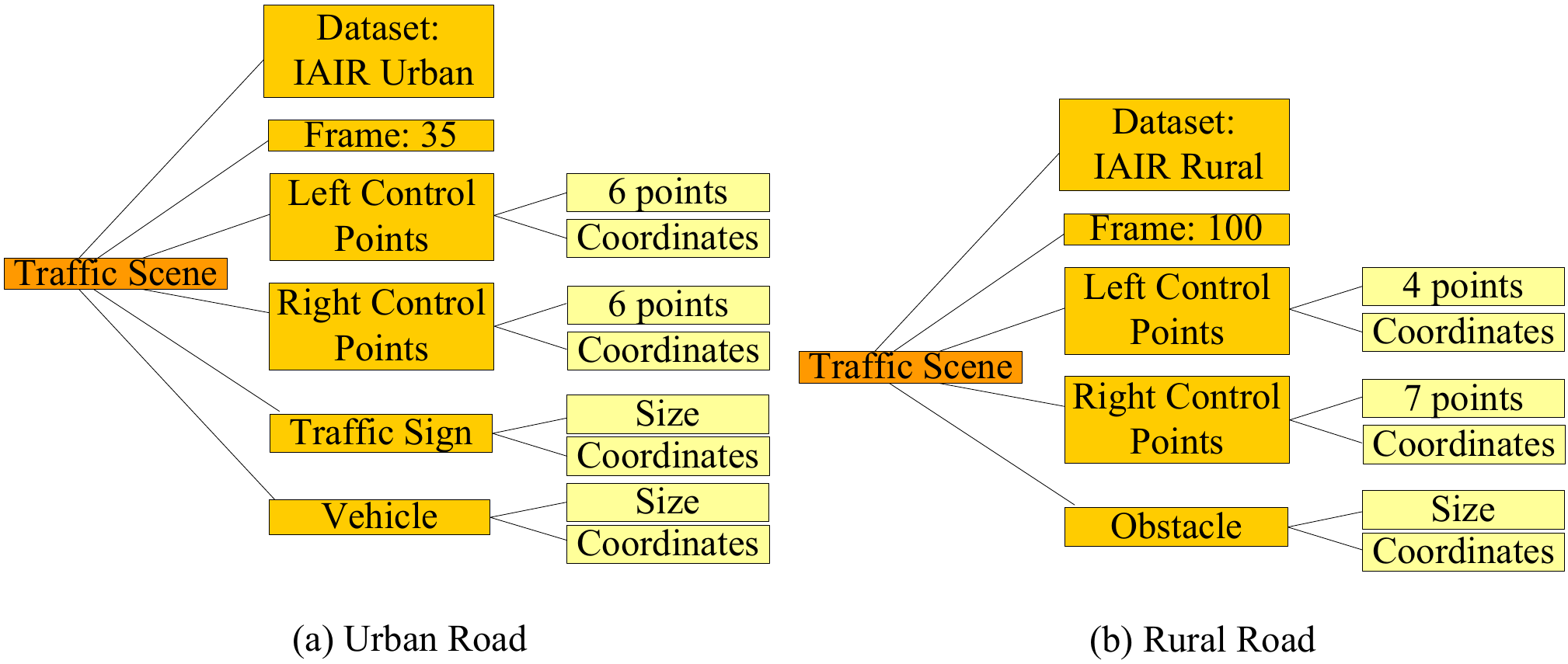}
\caption{Scene model database of the TSD-max dataset. (a) Urban road. (b) Rural road.}
\label{fig:Scene_Database}
\end{figure}

The background models for the 3D corridor structure can be constructed on the basis of the detection results for the road regions.
%Based on the detection results of road regions, the background models with the 3D Corridor structure are constructed.
The road region is set as a horizontal plane and, the rest of the background regions are assumed to stand perpendicularly to the road plane.
If there are any  foreground objects in the input image, a foreground/background segmentation method is adopted to construct the foreground and background models independently \cite{IS_Chen}.
%The foreground segmentation is implemented by a level set method without solving partial differential equations \cite{SPIC}.
The foreground models are  constructed with an $RGBA$ data structure, where $A$ denotes the transparency ratio.
The background inpainting is implemented by a process of ``optical flow inpainting/image inpainting''.
According to the 3D scene models, new viewpoint images can be generated by changing the views using OpenGL hardware texture mapping  \cite{T_ITS}, as shown in Fig. \ref{fig:new_viewpoint}.
%The new viewpoint images can be generated baded on the 3D scene models
In order to properly structure the 3D corridor models, a scene model database is constructed.
Each scene model corresponds to a frame in the image sequence, and a table to represent the connections  between the foreground and background is defined \cite{Building_Database}, as shown in Fig. \ref{fig:Scene_Database}.
The algorithm for road scene models construction is shown in Algorithm \ref{Alg_Scene}.
%As shown in Fig. \ref{fig:Scene_Database}, program administration such as new viewpoint rendering, scene simulation is implemented using the model database.

%The camera orientation is defined by three Euler angles $\alpha$ , $\beta$ and $\gamma$ in the Manhattan axis system.

%\subsubsection{Scene Reconstruction for Panoramic Images}
As well as the scene reconstruction for monocular images, panoramic images can  be used for scene construction.
The image stitching algorithm we used largely relates to global alignment and image blending \cite{Seamless_Stitch}.
In the global alignment stage, bundle adjustment, parallax removal and feature-based alignment are applied to roughly stitch together the input images.
Out of a wide choice of possible projection models, a cylindrical projection model is especially effective.
Other, de-ghosting and blending steps also need to be undertaken to create the output panoramas.
After this, the panoramic scene models can be constructed.
%Based on the panoramic scene models, we can further apply crossed-slits projection \cite{Crossed_Slits} to slightly change the positions of  objects on roadsides, e.g. buildings and trees.\par

%    \begin{figure}[!htbp]
%   \vspace{1ex}
%   \centering
%   \vspace{1ex}
%   \includegraphics[width=8.5cm]{figures/Image_Stitch.eps}
%   \caption{Image stitching process.}
%   \label{fig:Image_Stitching}
%   \vspace{1ex}
%   \end{figure}

%\subsubsection{Non-Photorealistic Scene Reconstruction}

%Besides the traditional rendering method, we apply non-photorealistic rendering to generate cartoon traffic images as a choice for users.
%The cartoon images are produced by the methods described in \cite{Artistic_Rendering}\cite{Circular_Threshold}.
%The global structures of the input images are preserved and enhanced.
%The rendering process follows a pipeline of `line detection-line selection-constrained segmentation-region simplification-rendering'.
%Circular thresholding is applied to process the color and texture features.

%\begin{figure}[!htbp]
%\centering
%\includegraphics[width=7.5cm]{figures/non_photorealistic.eps}
%\caption{Non-photorealistic rendering results.}
%\label{fig:non_photorealistic}
%\end{figure}

\section{Experiments and Applications}
The platform used for our experiments was a computer with an Intel i5 processor @3.33GHZ and 16.0 GB RAM.
The experiments for road region detection were implemented using MATLAB R14, while the scene construction experiments were based on the OpenGL Toolbox.
For the road detection, we designed experiments using three datasets: the Bristol Dataset \cite{Recognize_Traffic_Signs} ($272\times272$ pixel size, 500 frames); the Caltech Dataset \cite{Detection_Lane_Marker} ($320\times240$ pixel size, 1000 frames);  and the TSD-max Dataset \footnote{http://trafficdata.xjtu.edu.cn/index.do} ($256\times256$ pixel size, 2000 frames).
The experiments for the scene model construction were mainly based on the TSD-max Dataset, which was constructed by the Institute of Artificial Intelligence and Robotics at Xi'an Jiaotong University.
TSD-max  is the basic dataset for the ``Future Challenge'', a national annual competition for unmanned vehicles supported by the National Natural Science Foundation of China.
%The image capturing frequency is 1 HZ, and the vehicle to capture the images drives at the speed of 1m/s.\par

\subsection{Road Detection Experiment}

The proposed road detection method can be implemented for both  single images (SMRF1) and image sequences (SMRF2).
To quantitatively evaluate the road detection results, we apply the  precision ($Pre$) and recall ($Rec$) metrics to compare them with ground truth road regions.
Precision denotes the ratio of correct pixels to the detected road region, while recall denotes the ratio of the correct pixels to the benchmark road region.
In our experiments, precision and recall are defined as follows:
\begin{eqnarray}
Pre = \frac{{\left| {R \cap {R_G}} \right|}}{R},{\kern 1pt} {\kern 1pt} {\kern 1pt} {\kern 1pt} {\kern 1pt} {\kern 1pt} Rec = \frac{{\left| {R \cap {R_G}} \right|}}{{{R_G}}}
\end{eqnarray}
where $R$ and $R_G$ denote the detected region and the ground truth, respectively. 
The $F_\beta$ score can be computed by combining $Pre$ and $Rec$ as follows:
\begin{equation}
{F_\beta } = (1 + {\beta^2 })\frac{{Pre \cdot Rec}}{{({\beta^2 } \cdot Pre) + Rec}}
\end{equation}
The recall metric is emphasized if $\beta>1$, while the precision is emphasized if $\beta<1$.
%where $$
%We set $\alpha=0.5$ to treat precision and recall equally.
%The F-measure is applied to uniformly evaluate the detection results.

\begin{figure}[!htbp]
\centering
\includegraphics[width=14cm]{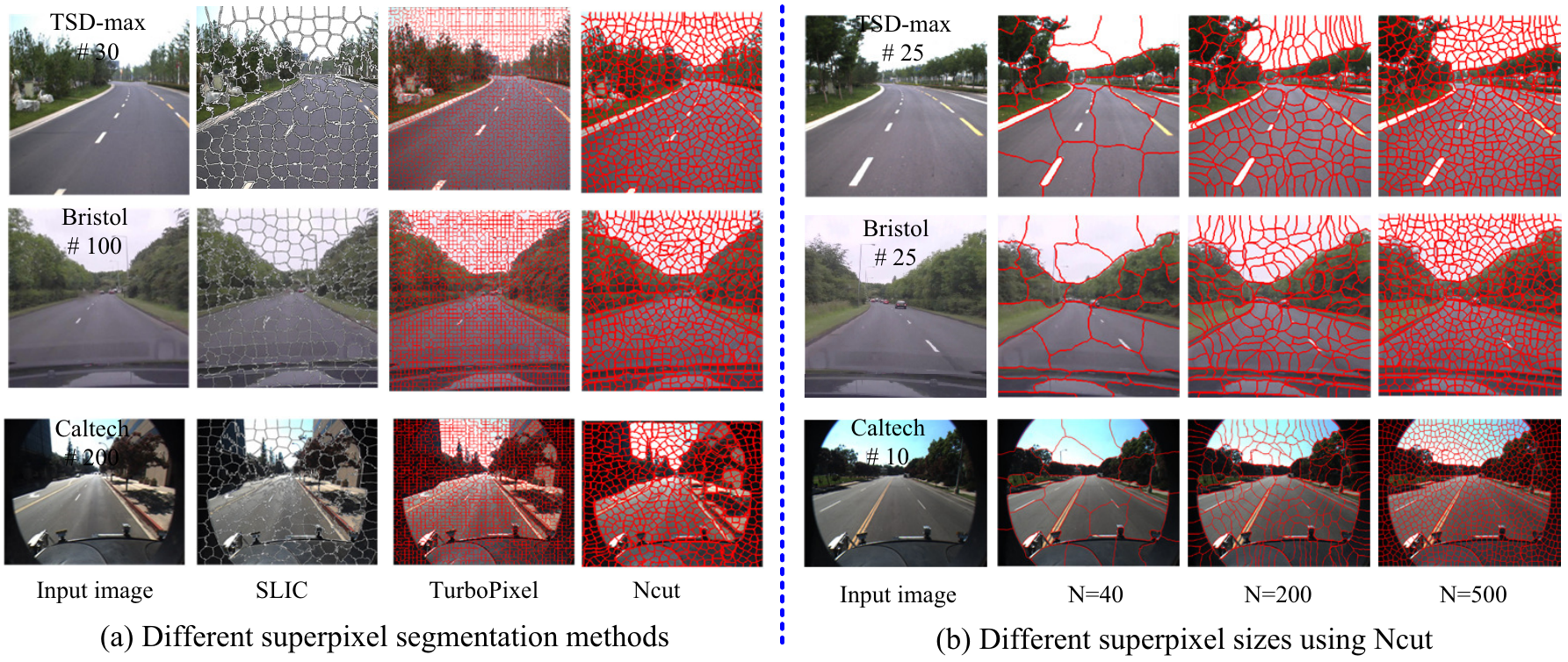}
\caption{Superpixel segmentation methods. (a) Different superpixel segmentation methods. (b) Different superpixel sizes using Ncut.}
\label{fig:superpixel_comparison}
\end{figure}

%\subsubsection{Superpixel size}
In the proposed method, superpixel segmentation is an important preliminary step.
%We apply Ncut method \cite{Ncut} to produce the uniformly segmented superpixels.
In Fig. \ref{fig:superpixel_comparison} (a), three superpixel segmentation methods are compared: the SLIC method, which is based on K-means clustering \cite{SLIC}; the Turbopixel method, which is based on multidimensional eigen-images \cite{TurboPixel}; and the Ncut method, which is based on graph partitioning \cite{Ncut}.
Superpixel segmentation can be implemented at different scales, as shown in Fig. \ref{fig:superpixel_comparison} (b).
Input images can be segmented using Ncut with the number of superpixels being set at $N=40$, $N=200$ and $N=500$.
For the sake of feature selection and convergence speed, we choose $N=500$ for the experiments.\par

   \begin{figure}[!htbp]
   \vspace{1ex}
   \centering
   \vspace{1ex}
   \includegraphics[width=9cm]{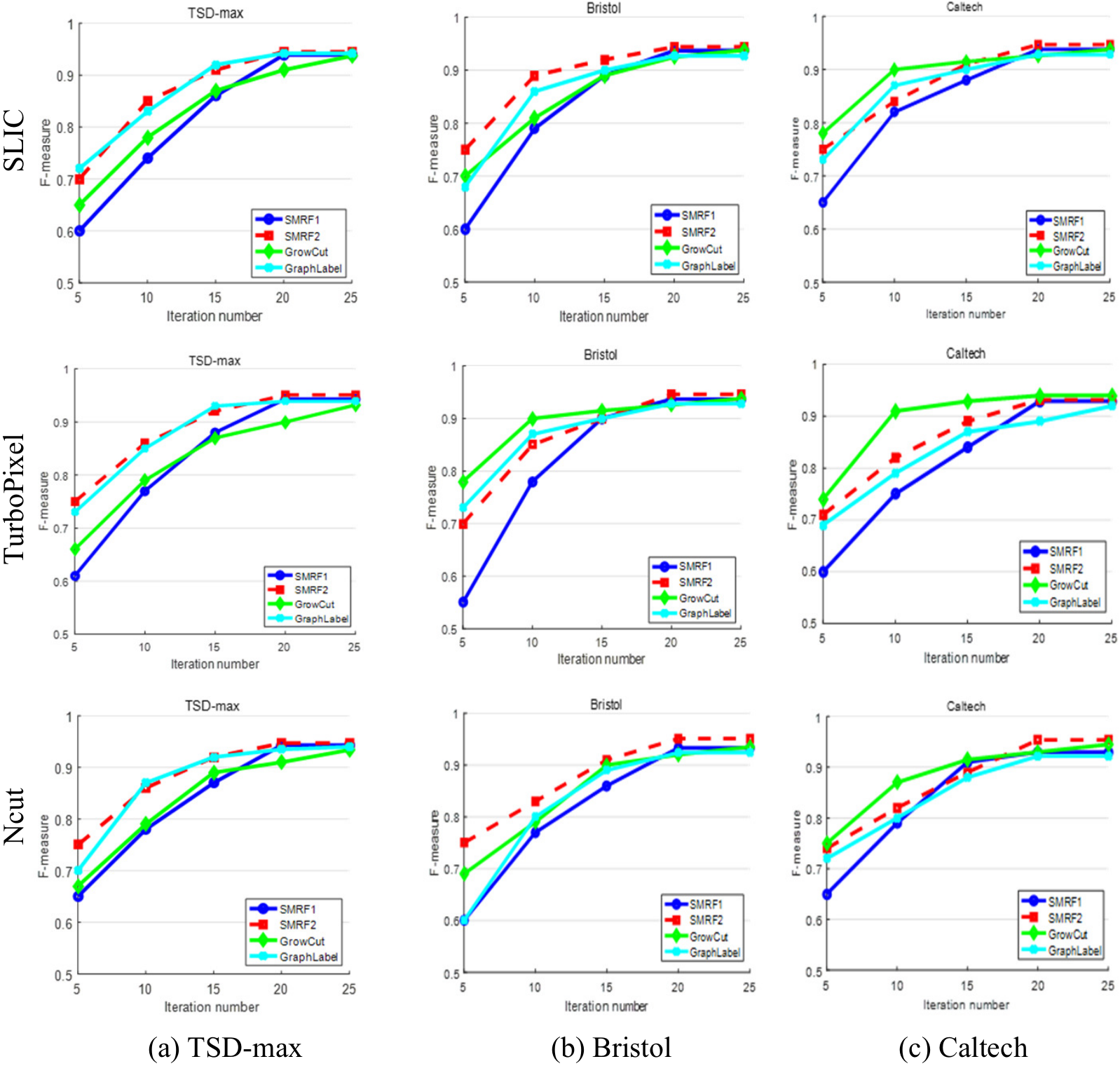}
   \caption{$F_\beta$ scores recorded at different algorithm iterations ($\beta=1$). (a) TSD-max Dataset. (b) Bristol Dataset. (c) Caltech Dataset.}
   \label{fig:iteration_numbers}
   \vspace{1ex}
   \end{figure}

\begin{table}[!htbp]
\centering
\caption{Number of iterations for convergence}
\begin{tabular}{c c c c c}
\hline
\textbf{} & \textbf{GrowCut} & \textbf{GraphLabel} & \textbf{SMRF1} & \textbf{SMRF2}\\
\hline
TSD-max  &25  &17  &18 &20\\

Bristol  &24 &17  &20 &20 \\

Caltech  &22 &18  &18 &20\\
%\hline
%Ours+Alvarez's\cite{road_detection_1} &93.0\% &87.5\%\\
\hline
\end{tabular}
\label{Tab:Iteration_Number}
\end{table}

After  specifying the number of superpixels, the influence of different numbers of iterations is evaluated.
Comparisons are undertaken between the proposed methods SMRF1 and SMRF2, two state-of-the-art methods GrowCut \cite{GrowCut} and GraphLabel \cite{Graph_Label}, using the TSD-max, Bristol and Caltech datasets.
The results based on different superpixel types are also compared.
We compute the average $F_\beta$ ($\beta=1$) score for every 5 iterations,  as depicted in Fig. \ref{fig:iteration_numbers}.
The comparison results indicate that, as the number of iterations increases, the $F_\beta$ score becomes more accurate.
However, peak values are reached after 20 iterations for most of the datasets.
Out of the three superpixel segmentation methods, Ncut obtains the most accurate results.
The iteration speed of the proposed method is similar to the Graph label method, and ourperforms the GrowCut method.
More detailed numbers of algorithm iterations are recorded in Table \ref{Tab:Iteration_Number}.

\begin{figure}[t]
\centering
\includegraphics[width=13cm]{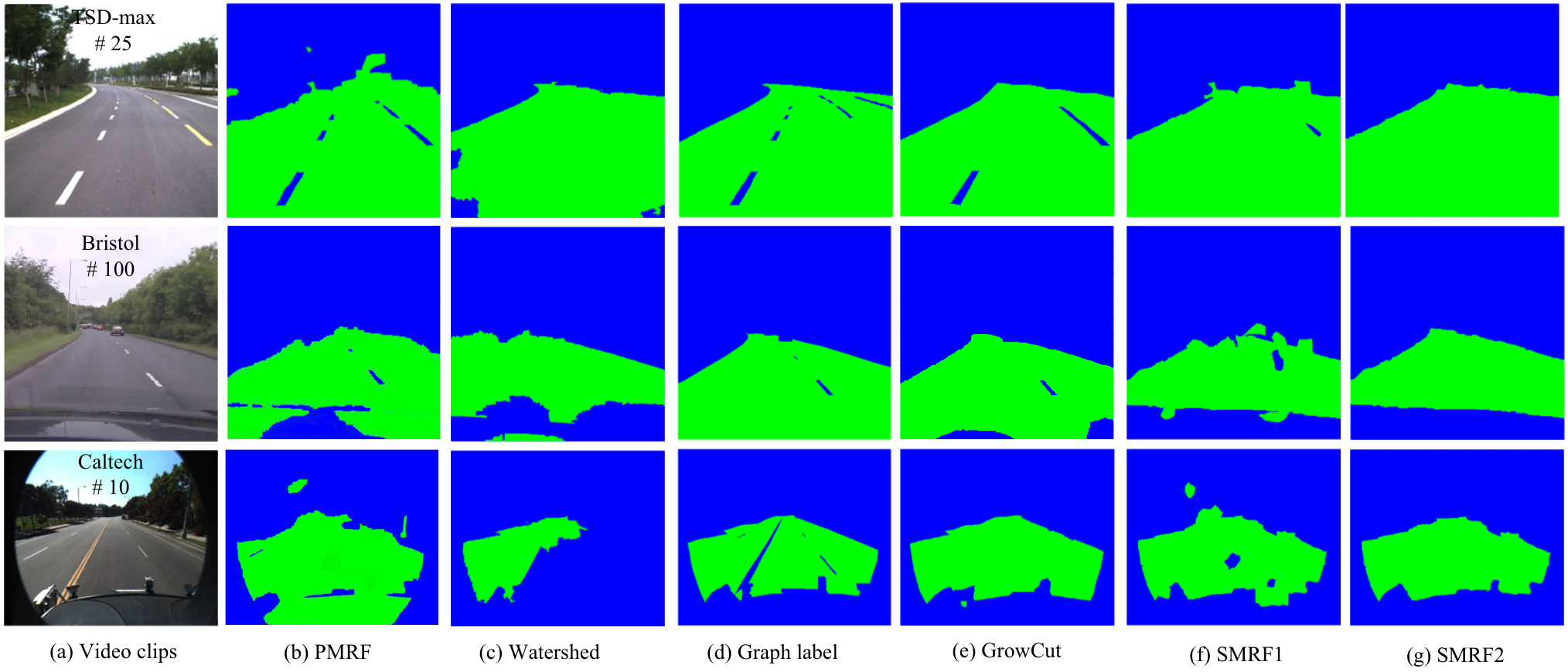}
\caption{Road region detection results. (a) Input image. (b) Pixel-based MRF \cite{Elia_CD}. (c) Watershed \cite{Watershed}. (d) Graph label \cite{Graph_Label}. (e) GrowCut \cite{GrowCut}. (f) SMRF1 (proposed). (g) SMRF2 (proposed).}
\label{fig:region_compare}
\end{figure}

Furthermore, a comparison is made between the proposed methods (SMRF1 and SMRF2) and state of the art methods.
% we evaluate the accuracy of the proposed methods by comparing with the baseline methods.
The pixel-based MRF \cite{Elia_CD} is implemented based on a tree-structured MRF model, which segment the input image into smaller regions following a ``split-merge'' strategy.
The Watershed method \cite{Watershed} uses a topological gradient approach to avoid over segmentation.
The Graph label method \cite{Graph_Label} is  based on label transfer and, uses an approximate nearest neighbor algorithm.
In the GrowCut framework \cite{GrowCut}, the superpixel-level seeds are selected using an unsupervised way, and the superpixel neighbors are detected.
The Superpixel-based conditional random field (SCRF) method \cite{Superpixel_CRF} regularizes a classifier by aggregating histograms in the superpixel neighborhoods, and apply a conditional random filed model for refinement.\par

\begin{table}[!htbp]
	\centering
	\caption{Quantitative evaluation of road detection results }
	\begin{tabular}{c l c c c c c}
		\hline
		{Datasets} & {Methods} & {Precision} & {Recall} & \multicolumn{3}{c} {$F_\beta$ score} \\
		& & & & {$\beta$=0.5} &{$\beta$=1} &{$\beta$=2} \\
		\hline
		
		\multirow{3}{*}\textbf{TSD-max}
        &  PMRF \cite{Elia_CD}                        & 0.8625 &0.8830  &0.8665 &0.8726 &0.8788 \\
		& Watershed \cite{Watershed}                     &0.8870   &0.8920 &0.8880 &0.8895 &0.8910 \\
		& Graph label \cite{Graph_Label}                   & 0.9410 &0.9425  &0.9423 &0.9442 &0.9462 \\
		& GrowCut \cite{GrowCut}                       & 0.9310 &0.9425  &0.9333 &0.9367 &0.9402 \\
        & SCRF \cite{Superpixel_CRF}                     &0.7680   &0.7920 &0.7727 &0.7798 &0.7871 \\
		& SMRF1+S1                      & 0.9320 &0.9370  &0.9330 &0.9345 &0.9360 \\
		& SMRF1+S2                      & 0.9380 &0.9290  &0.9362 &0.9335 &0.9308 \\
		& SMRF1+S3                      & 0.9385 &0.9325  &0.9373 &0.9355 &0.9337 \\
		& SMRF2+S1                      & \bf{0.9465} &\bf{0.9550}  &\bf{0.9482} &\bf{0.9507} &\bf{0.9533} \\
		& SMRF2+S2                      & 0.9410 &0.9495  &0.9427 &0.9452 &0.9478 \\
		& SMRF2+S3                      & 0.9405 &0.9535  &0.9431 &0.9470 &0.9509 \\
        \hline
		\multirow{3}{*}\textbf{Bristol}
        &  PMRF \cite{Elia_CD}                          & 0.8425 &0.8320  &0.8404 &0.8372 &0.8341 \\
		& Watershed \cite{Watershed}                     & 0.8375 &0.8565  &0.8412 &0.8469 &0.8526 \\
		& Graph label \cite{Graph_Label}                   & 0.9250 &0.9305  &0.9261 &0.9277 &0.9294 \\
		& GrowCut \cite{GrowCut}                       & 0.9350 &0.9410  &0.9362 &0.9380 &0.9398 \\
        & SCRF \cite{Superpixel_CRF}                      &0.8250  &0.8275 &0.8255 &0.8262 &0.8270 \\
		& SMRF1+S1                      & 0.9295 &0.9340  &0.9304 &0.9317 &0.9331 \\
		& SMRF1+S2                      & 0.9350 &0.9400  &0.9360 &0.9375 &0.9390 \\
		& SMRF1+S3                      & 0.9345 &0.9315  &0.9339 &0.9330 &0.9321 \\
		& SMRF2+S1                      & 0.9400 &0.9485  &0.9417 &0.9442 &0.9468 \\
		& SMRF2+S2                      & 0.9360 &\bf{0.9580}  &0.9403 &0.9469 &0.9535 \\
		& SMRF2+S3                      & \bf{0.9480} &0.9552  &\bf{0.9494} &\bf{0.9516} &\bf{0.9538} \\

        \hline
		\multirow{3}{*}\textbf{Caltech}
        &  PMRF \cite{Elia_CD}                          & 0.8210 &0.7945  &0.8156 &0.8075 &0.7997 \\
		& Watershed \cite{Watershed}                     & 0.6250 &0.6500  &0.6298 &0.6373 &0.6448 \\
		& Graph label \cite{Graph_Label}                   & 0.9320 &0.9200  &0.9296 &0.9260 &0.9224 \\
		& GrowCut \cite{GrowCut}                       & 0.9500 &0.9480  &0.9496 &0.9490 &0.9484 \\
        & SCRF \cite{Superpixel_CRF}                      &0.8185  &0.8310 &0.8210 &0.8247 &0.8285 \\
		& SMRF1+S1                      & 0.9290 &0.9410  &0.9314 &0.9350 &0.9385 \\
		& SMRF1+S2                      & 0.9220 &0.9370  &0.9250 &0.9294 &0.9340 \\
		& SMRF1+S3                      & 0.9325 &0.9280  &0.9316 &0.9302 &0.9289 \\
		& SMRF2+S1                      & 0.9420 &0.9485  &0.9433 &0.9452 &0.9472 \\
		& SMRF2+S2                      & 0.9380 &0.9270  &0.9350 &0.9320 &0.9290 \\
		& SMRF2+S3                      & \bf{0.9560} &\bf{0.9520}  &\bf{0.9552} &\bf{0.9540} &\bf{0.9528} \\
        \hline
		
	\end{tabular}
\label{tab:road_detection}
\end{table}

The detection results based on these state-of-the-art methods are depicted in Fig. \ref{fig:region_compare}, with green and blue  denoting the detected road regions and the non-road regions, respectively.
The  results prove that the proposed method (SMRF2) achieves the most accurate detection results from the point-of-view of visual aesthetics.
%In the datasets of Bristol and Caltech, a part of vehicle exists in the image bottom area.
The other methods perform less well than our own, with the   vehicle and  road regions appearing too similar.
The reason that PMRF and Watershed methods perform worse than the proposed methods is that they lack the statistical features of superpixels.
The Graph label and GrowCut methods ignore the sptio-temporal  interactions between adjacent superpixels.\par

Aside from the qualitative comparisons,  evaluations are undertaken on the basis of the $F_\beta$ score, as summarized in Table \ref{tab:road_detection}.
The parameter  $\beta$ is variously set to 0.5, 1 and 2.
As mentioned above, the detection precision is emphasized when $\beta>1$, while the recall is emphasized when $\beta<1$.
The state-of-the-art methods are again implemented for comparison.
%SMRF2 is evaluated for our method.
We compute the average $F_\beta$ score over all of the complete image sequences for each dataset.
The comparative results show that the $F_\beta$ scores for SMRF2 are more than 0.927 for all  3 datasets, outperforming PMRF, Watershed, Graph label and Grow cut.
This result is consistent with the qualitative results shown in Fig. \ref{fig:region_compare}.
We also evaluate the results for different superpixel segmentations, with S1, S2 and S3 denoting the segmentations for TurboPixel \cite{TurboPixel}, SLIC \cite{SLIC} and Ncut \cite{Ncut}, respectively.
The other methods are also implementedusing the different superpixel types, however, we only record their best performances.
% the best results using different superpixel segmentations.
Comparison reveals that  the proposed SMRF2 method outperformes all of the other methods.
The results of  SMRF1  are similar to the Graph label method.
Among the three superpixel segmentation types, the highest $F_\beta$ are reached by the Ncut method.
\par

   \begin{figure}[!htbp]
   \vspace{1ex}
   \centering
   \vspace{1ex}
   \includegraphics[width=8cm]{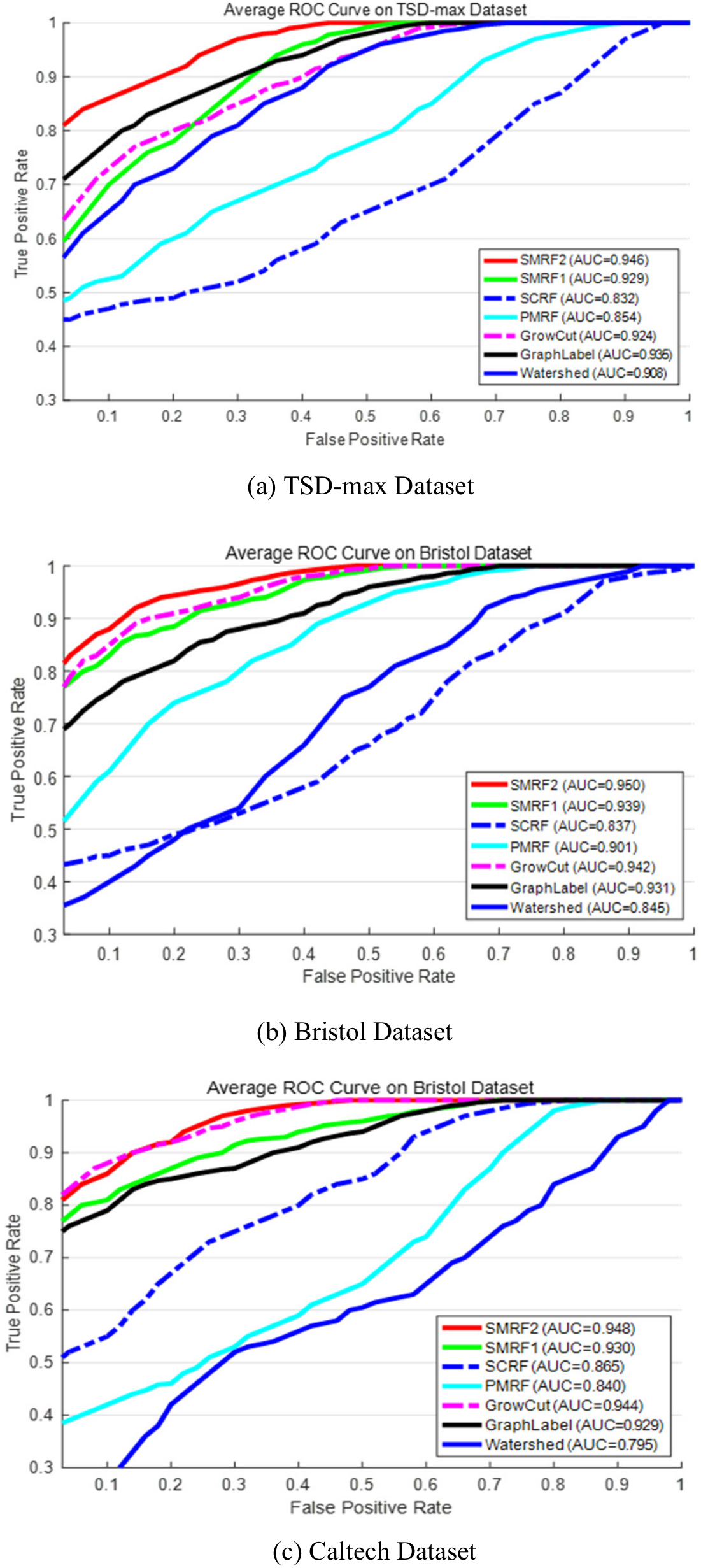}
   \caption{Average ROC Curves. (a) TSD-max Dataset. (b) Bristol Dataset. (c) Caltech Dataset.}
   \label{fig:ROC_Curve}
   \vspace{1ex}
   \end{figure}

\begin{figure}[t]
   \vspace{1ex}
   \centering
   \vspace{1ex}
   \includegraphics[width=11cm]{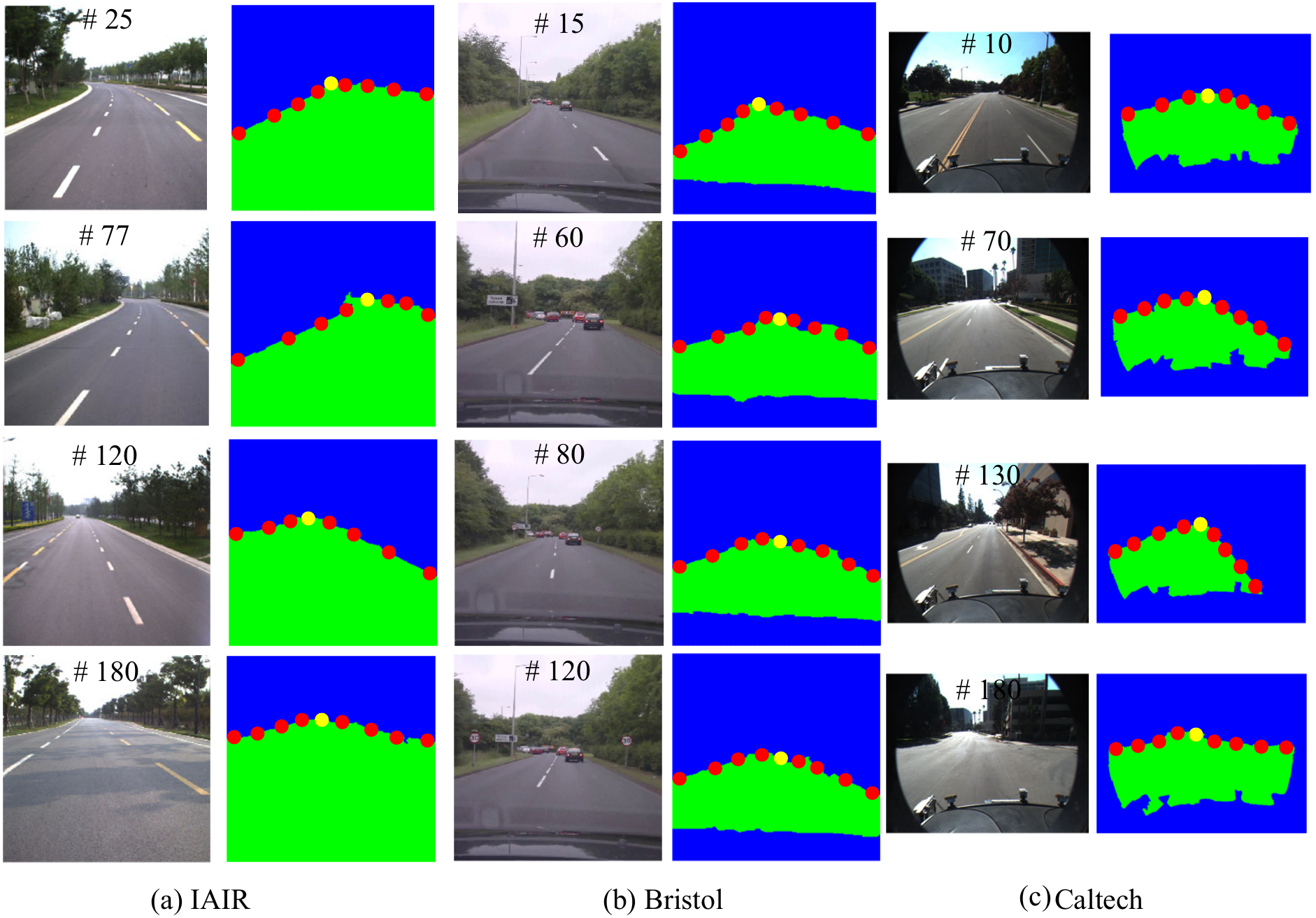}
   \caption{Experimental results of SMRF2. (a) TSD-max dataset. (b) Bristol dataset. (c) Caltech dataset.}
   \label{fig:Results_SMRF2}
   \vspace{1ex}
   \end{figure}

Moreover, we compare the average receiver operating characteristic (ROC) curves on the pixelwise comparison between the detected road regions and the ground truth.
ROC curves denote the tradeoff between the true positive rate $TPR=(TP/TP+FN)$ and the false positive rate $FPR=(FP/FP+TN)$.
$TP$ and $TN$ are the number of road pixels correctly detected and the background pixels correctly deteccted, respectively.
$FP$ and $FN$ are the number of background pixels incorrectly marked and the road pixels incorrectly identified, respectively.
The  area under curve (AUC) metric can be easily evaluated based on the ROC curves.
As the comparison results show, the highest AUC value is provided by the proposed SMRF2 method for all the datasets; see Fig. \ref{fig:ROC_Curve}.
The performances of SMRF2 and GrowCut method are similar based on the Caltech dataset.
The performances of the SMRF1 method are better than PMRF and Watershed methods.
Further detection results for SMRF2 are depicted in Fig. \ref{fig:Results_SMRF2}, along with the control points and vanishing points.\par

%After the detection of road regions, the corresponding road contours can be generated.
%Besides the comparison of road region, the accuracy of road contour is used as another evaluation metric.
%We compare the proposed method (SMRF2) with the methods of Alvarez \cite{Alvarez} and Yao \cite{Yao_J}.
%The average tracking error (ATE) is applied to the evaluation process.
%For each pixel on the detected contour of $i_{th}$ frame, we find the corresponding nearest pixel on the benchmark road contour.
%The distance of the pixels are accumulated as the tracking error of $i_{th}$ frame ($TE_i$).
%The total tracking error (TTE) of an image sequence with $N$ frames is defined as: $\mbox{TTE}=\sum_{i=1}^{N}\mbox{TE}_i$.
%The average tracking error is defined as $\mbox{ATE}=\mbox{TTE}/\mbox{N}$.
%The comparisons of road contours are shown in Fig. \ref{fig:Compare_Road_Contour}, where the tracking error of every 10 frames is depicted.
%Tab. \ref{tab:comparison_road_contour} summarizes the ATE based on the mentioned three datasets.
%As the comparison results show, the SMRF2 method achieves more accurate tracking results compared to Alvarez and Yao methods.
%More road detection results with control points generation based on SMRF2 are shown in Fig. \ref{fig:Results_SMRF2}.

\subsection{Scene construction Experiment}

   \begin{figure}[t]
   \vspace{1ex}
   \centering
   \vspace{1ex}
   \includegraphics[width=14cm]{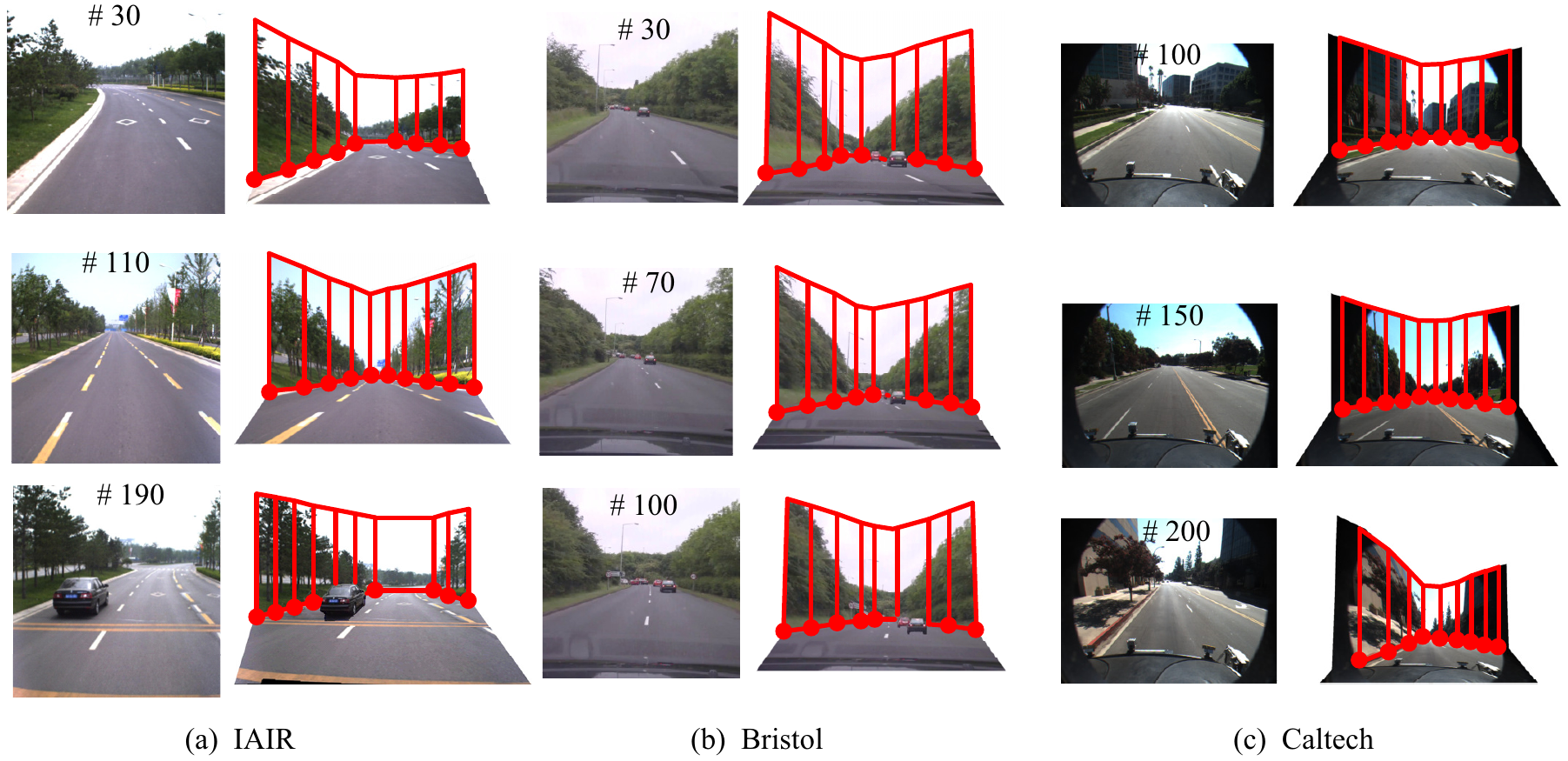}
   \caption{Scene models construction results. (a) TSD-max dataset. (b) Bristol dataset. (c) Caltech dataset.}
   \label{fig:Scene_Models}
   \vspace{1ex}
   \end{figure}

\begin{figure}[!htbp]
\centering
\includegraphics[width=14cm]{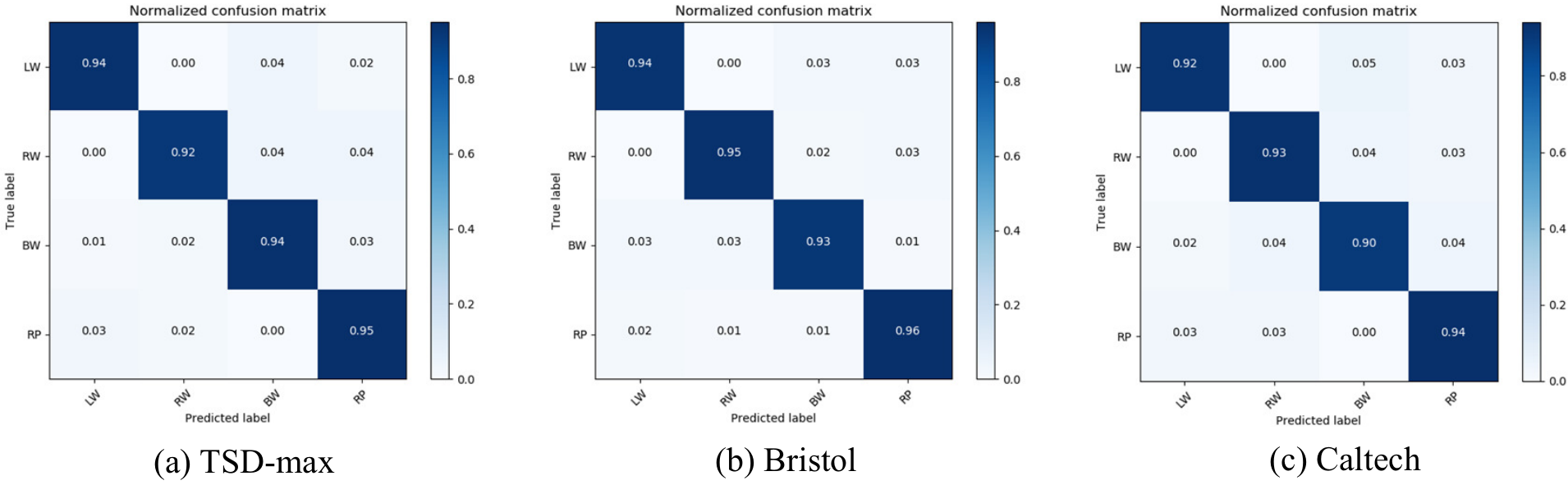}
\caption{Confusion matrix for the scene model components. (a) TSD-max. (b) Britol. (c) Caltech.}
\label{fig:confusion_matrix}
\end{figure}

Based on the road detection results, we evaluate the accuracy and effectiveness of the road scene models.
Scene models based on the different datasets are shown in Fig. \ref{fig:Scene_Models}.
We used a confusion matrix as a metric to evaluate the accuracy of the  scene model components in pixels, with LW, RW, BW and RP denoting ``left wall'', ``right wall'', ``back wall'' and ``road plane'', as shown in Fig. \ref{fig:confusion_matrix}.
The ground truth for scene models are annotated in advance.
The evaluation results indicate the correctness ratios of the proposed scene models are beyond 0.9 for these scene components.\par

\begin{figure}[!htbp]
   \vspace{1ex}
   \centering
   \vspace{1ex}
   \includegraphics[width=13.5cm]{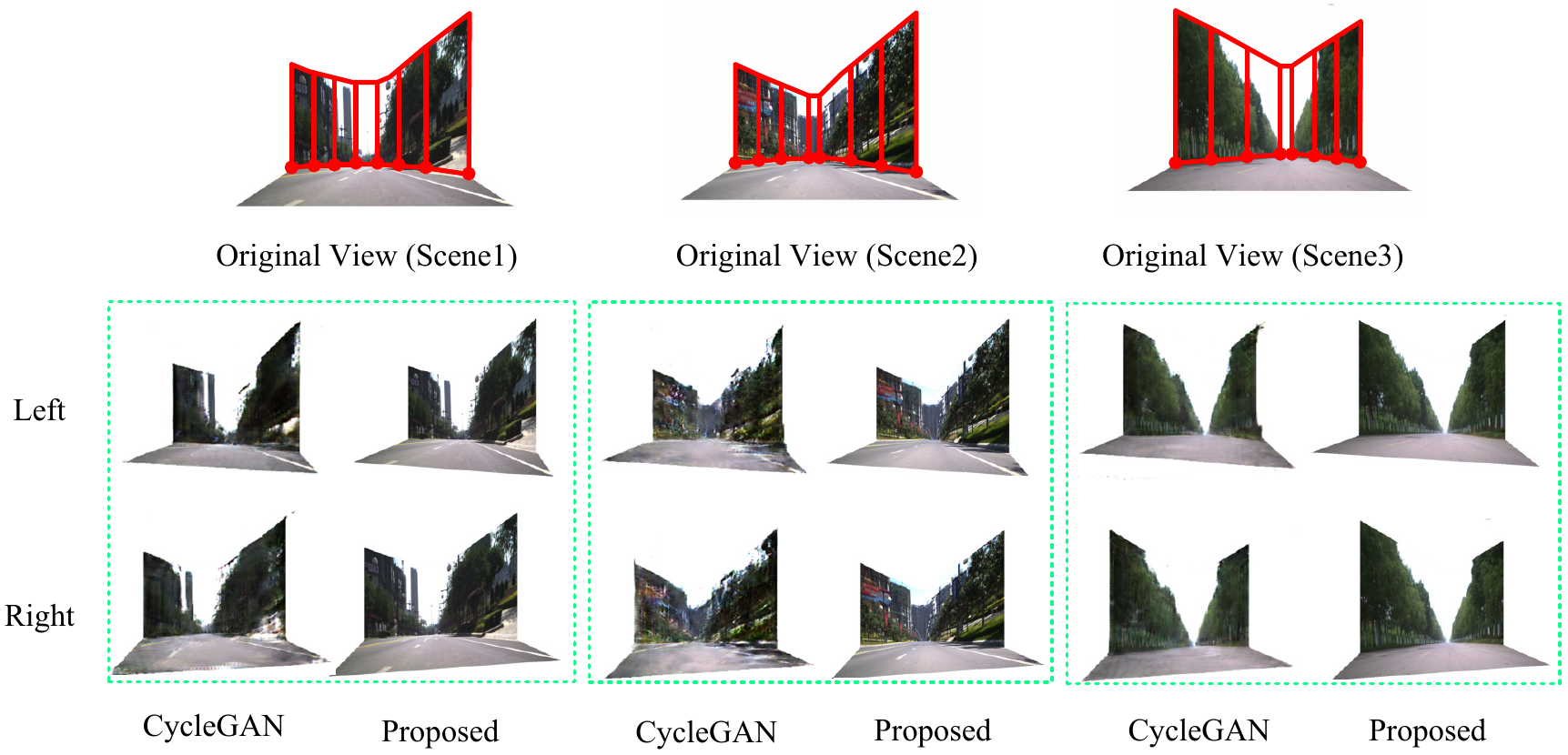}
   \caption{Experiments for view angle change.}
   \label{fig:View_Compare}
   \vspace{1ex}
   \end{figure}

As we change the positions and angles of the scene models, new viewpoint images are generated by projecting the scene models onto the output image planes, as exhibited in Fig. \ref{fig:new_viewpoint}. 
To assess the stability of the proposed scene models, we compare them to different view angles generated using CycleGAN \cite{CycleGAN}.
%, as shown in Fig. \ref{fig:View_Compare}.
The CycleGAN network consists of 2 generators and 2 discriminators.
Images are generated based on the generators and discriminators at different view angles.
%The training data of the CycleGAN network include the original images and the new images under left and right view angles.
Different view angle images are compared in Fig. \ref{fig:View_Compare}.
The results show that, compared to the proposed method, the textures of different view angle images using CycleGAN are fuzzy, and prone to distortion.\par

In order to quantitatively evaluate the proposed scene models, we apply two metrics \cite{Saxena}:
(1) The plane correctness ratio, where a plane is defined as correct if more than 75\% of its patches are correctly detected as semantic wall and road regions.
(2) The model correctness ratio, where a model is defined as correct if more than 75\% of its patches in the wall and road planes have a correct relationship to their neighbors.
The ratio of plane patches to texture distortions is less than 25\%.\par

   \begin{figure}[!htbp]
   \vspace{1ex}
   \centering
   \vspace{1ex}
   \includegraphics[width=13cm]{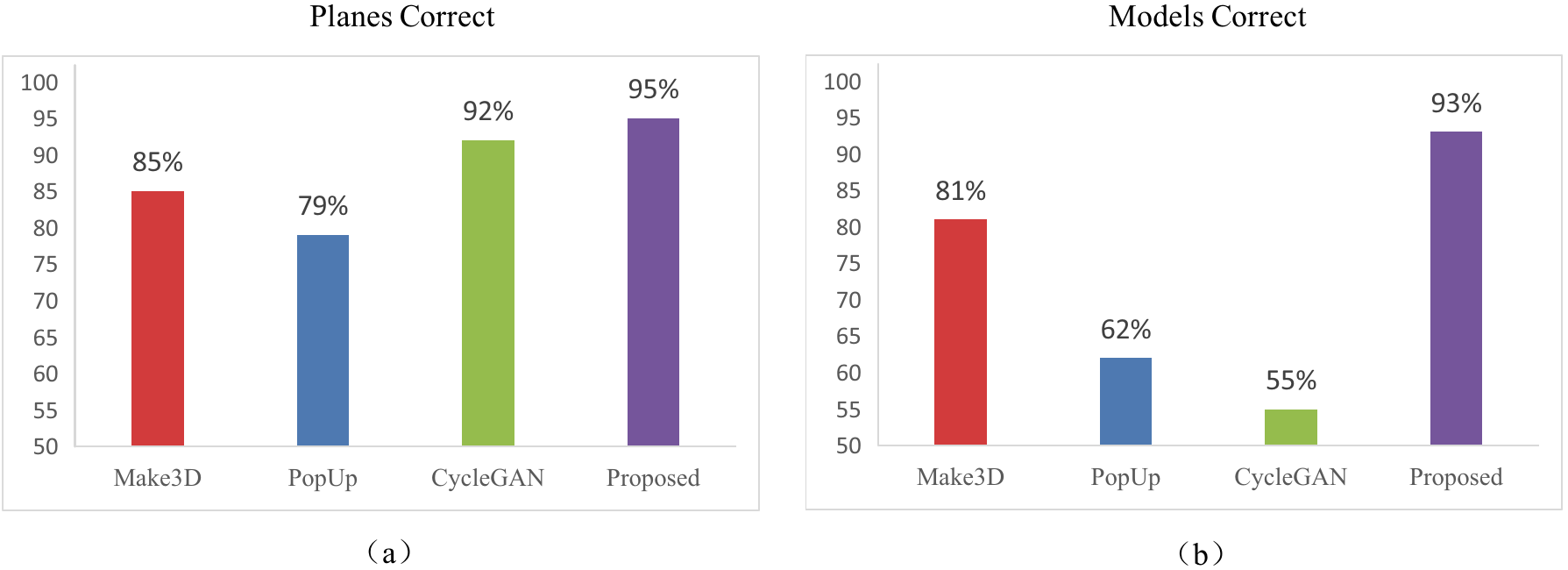}
   \caption{Models comparison. (a) Plane correctness ratio. (b) Model correctness ratio.}
   \label{fig:plane_correctness}
   \vspace{1ex}
   \end{figure}

The evaluation was conducted by a person not associated with the project, who applied the above metrics.
1000 images from the TSD-max dataset are chosen for the experiment.
Under these metrics, we compare the proposed models by Make3D \cite{Saxena}, PopUp \cite{Hoiem} and CycleGAN \cite{CycleGAN}, as shown in Fig. \ref{fig:plane_correctness}.
%The plane and model correctness ratio prove the effectiveness of our method.
The  results confirm that our models achieve both the best plane correctness ratio and the best model correctness ratio.
The plane correctness ratio of CycleGAN method is close to us, however, its model correctness ratio  is far inferior.\par
%More scene construction results are shown in Fig. \ref{fig:Scene_Models}, which are based on the TSD-max, Bristol and Caltech datasets.\par

%Based on the scene models construction results, useful applications can be developed.\par

\subsection{Applications}

\begin{figure}[!htbp]
\vspace{1ex}
\centering
\vspace{1ex}
\includegraphics[width=12cm]{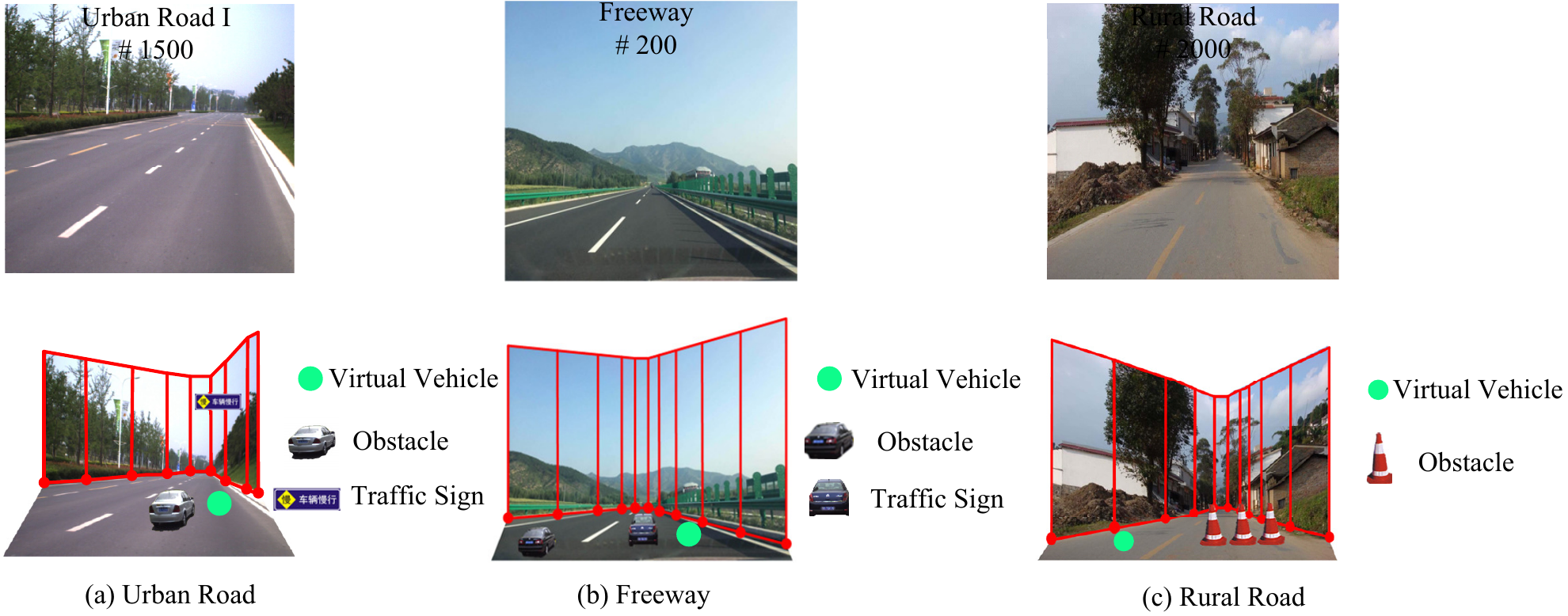}
\caption{Scene simulation examples. (a) Urban road. (b) Freeway. (c) Rural road.}
\label{fig:Scene_Simulation}
\vspace{1ex}
\end{figure}

   \begin{figure}[!htbp]
   \vspace{1ex}
   \centering
   \vspace{1ex}
   \includegraphics[width=10cm]{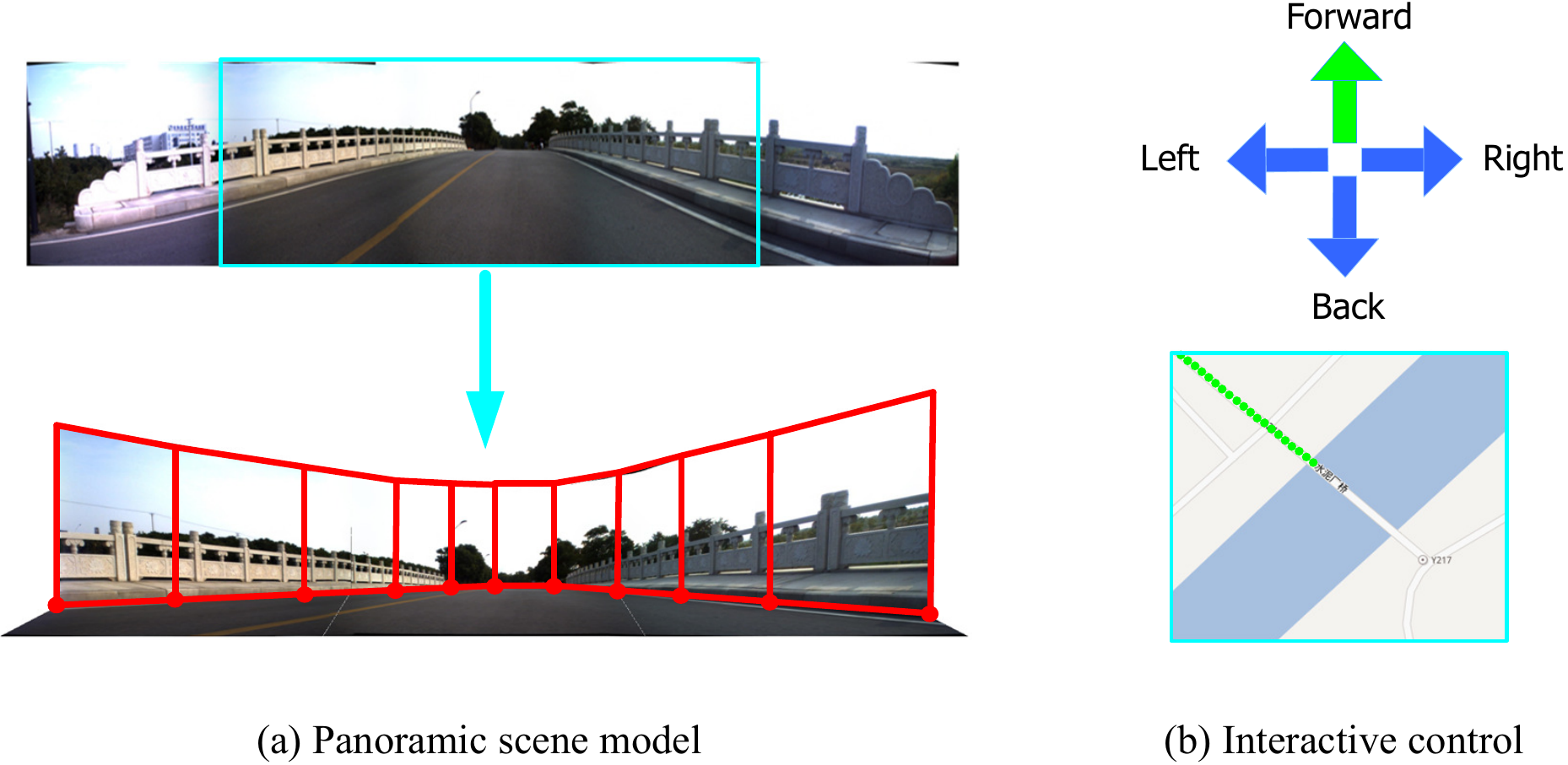}
   \caption{Virtual street view. (a) Panoramic scene models. (b) Interactive control and trajectory rendering.}
   \label{fig:Scene_Panorama}
   \vspace{1ex}
   \end{figure}

%   \begin{figure}[!htbp]
%   \vspace{1ex}
%   \centering
%   \vspace{1ex}
%   \includegraphics[width=8.5cm]{figures/Map_trajectory2.pdf}
%   \caption{Street view based on panoramic scene models. (a) Panoramic scene models and street view. (b) Interactive control and trajectory rendering.}
%   \label{fig:Scene_Panorama}
%   \vspace{1ex}
%   \end{figure}

Various applications focused on traffic scene simulation can be developed based on the proposed scene models, as shown in Fig. \ref{fig:Scene_Simulation}.
The composite system we propose here is named  `Tour Into the Traffic Video' (TITV).
This can offer two simulation modes: (1) a bird's-eye-view mode; and (2) a touring mode.
%   The position, direction and speed of the virtual vehicle is rendered and controlled by the users.
   In the bird's-eye-view mode, the position, direction and speed of the virtual vehicle can be controlled by the user.
   Additional objects can be supplemented into the traffic scenes, such as obstacles, traffic signs, tower beacons, etc.
   Users can also observe the simulation from a  bird's-eye point-of-view.
%   The directions and speed of the virtual vehicle can be controlled.
   In the touring mode, new viewpoint images can be rendered as the viewpoint moves; see Fig. 4(b).
   The touring mode provides users with a  feeling of actually being on the road.\par

   Panoramic scene models can also be constructed for the virtual street view, as shown in Fig. \ref{fig:Scene_Panorama}.
   Here, users can tour the street scenes using the commands  ``move forward'', ``move back'', ``turn left'' and ``turn right'', which are consistent with the actions of a real driver.
%   The new viewpoint images are generated accordingly.
   At the same time, the tour trajectory is displayed on a map using GPS data.\par

In Table \ref{Tab:System_Compare}, the functionalities of the proposed system are compared with those offered by Google Street View (GSV) \cite{Google_View} and Microsoft Street Slide (MSS) \cite{Street_Slide}.
%$\surd$ means the system holds the corresponding functionality, while $\times$ means the system lacks the functionality.
The results emphasize the novelty of our system, with it providing:
(1) the scope to incorporate panoramas;
(2) independent modeling of foreground objects;
(3) generation of free viewpoint images;
(4)  display of map trajectories;
%(4) cartoon images can be rendered.

\begin{table}[!htbp]
\centering
\caption{Scene functionalities comparison}
\begin{tabular}{c c c c c}
\hline
\textbf{} & \textbf{Panorama} & \textbf{Foreground} & \textbf{Free Viewpoint} & \textbf{Map}\\
\hline
GSV \cite{Google_View} &Y &N  &N &Y\\

MSS \cite{Street_Slide} &Y &N  &N &Y \\

TITV &Y &Y  &Y &Y\\
%\hline
%Ours+Alvarez's\cite{road_detection_1} &93.0\% &87.5\%\\
\hline
\end{tabular}
\label{Tab:System_Compare}
\end{table}

\section{Conclusion and Future Works}
In this paper, we have proposed a new framework for constructing spatio-temporal scene models from road image sequences.
The reconstructed scene models have a 3D corridor structure, with road region detection being a precondition for scene construction.
We have developed a new superpixel-based MRF method for road detection, which follows a cycle of ``global energy initialization/local energy computation/global energy comparison''.
The data fidelity term of the energy function is defined according to a combination of color, texture and location features.
For both single images and image sequences, the smoothness term of the energy function is defined on the basis of superpixel interactions with spatially and spatio-temporally neighboring superpixels, respectively.\par

On the basis of the road region detection results, road boundary control points are generated to construct the scene models.
The scene models have a 3D corridor structure, with the road regions being assumed to be floor planes.
Panoramic scene models can be constructed to offer users more choices.
Applications for the simulation of unmanned vehicles have also been developed.\par

In our future work, a depth map will be utilized as a supplement for scene reconstruction.
On the basis of this,  more detailed scene surface constructions can be implemented.
We are also exploring ways to  design algorithms that can detect additional label types, such as buildings and pedestrians.

%\appendices

%\section{}

\section*{Acknowledgment}

This work is supported by National Natural Science Foundation of China under Grant No. 61803298, Natural Science Foundation of Jiangsu Province under Grant No. BK20180236, Suzhou Key Industry Technological Innovation (Perspective Application Research) under Grant No. SYG201843.
The authors thank the anonymous reviewers for their valuable comments and suggestions.
%We thank the anonymous reviewers for their valuable comments and suggestions.


\end{document}